\documentclass[lettersize,journal]{IEEEtran}
\usepackage{amsmath,amsfonts}
\usepackage{algorithmic}
\usepackage{algorithm}
\usepackage{array}
\usepackage{textcomp}
\usepackage{stfloats}
\usepackage{url}
\usepackage{verbatim}
\usepackage{graphicx}
\usepackage[noadjust]{cite}

\usepackage{amssymb}
\usepackage[pdfstartview=XYZ,
bookmarks=true,
colorlinks=true,
linkcolor=blue,
urlcolor=blue,
citecolor=blue,
pdftex,
bookmarks=true,
linktocpage=true, %
hyperindex=true
]{hyperref}

\usepackage{orcidlink}
\usepackage{booktabs}
\usepackage{multirow}
\usepackage{glossaries}

\usepackage{longtable}
\usepackage{amsmath,amsfonts}
\usepackage{algorithmic}
\usepackage{array}
\usepackage[caption=false,font=scriptsize,labelfont=sf,textfont=sf]{subfig}
\usepackage{textcomp}
\usepackage{stfloats}
\usepackage{url}
\usepackage{verbatim}
\usepackage{graphicx}
\usepackage{balance}

\newcommand{\AmountOfTrainingALCycles}{c}
\newcommand{\AmountOfTrainingALCyclesSet}{\mathbb{C}}

\newcommand{\unlabeledQuery}{Q}

\newcommand{\singleSample}{x}
\newcommand{\singleLabel}{y}
\newcommand{\amountOfSamples}{n}

\newcommand{\ALExperiment}{E}
\newcommand{\ALExperimentResult}{R}
\newcommand{\ALDataset}{D}
\newcommand{\ALTrainTestSplit}{\mathcal{T}}

\newcommand{\ALStartSet}{\mathcal{I}}
\newcommand{\labeledSet}{\ALDataset_{lab}}
\newcommand{\trainSet}{\ALDataset_{train}}
\newcommand{\testSet}{\ALDataset_{test}}
\newcommand{\unlabeledSet}{\ALDataset_{unl}}

\newcommand{\ALMetric}{M}

\newcommand{\learner}{\mathcal{L}}
\newcommand{\ALStrategy}{\mathcal{S}}
\newcommand{\batchSize}{b}
\newcommand{\pearson}{r}

\newcommand{\ALExperimentSet}{\mathbb{E}}
\newcommand{\ALExperimentResultsSet}{\mathbb{R}}
\newcommand{\ALDatasetSet}{\mathbb{D}}
\newcommand{\ALTrainTestSplitSet}{\mathbb{T}}
\newcommand{\ALStartSetSet}{\mathbb{I}}
\newcommand{\ALMetricSet}{\mathbb{M}}
\newcommand{\learnerSet}{\mathbb{L}}
\newcommand{\batchSizeSet}{\mathbb{B}}
\newcommand{\ALStrategySet}{\mathbb{S}}

\newcommand{\ALFingerprintVector}{V}
\newcommand{\leaderboardCell}{LC}
\newcommand{\finalRanking}{FR}

\newacronym{ML}{ML}{Machine Learning}
\newacronym{NLP}{NLP}{Natural Language Processing}
\newacronym{CNN}{CNN}{Convolutional Neural Networks}
\newacronym{AL}{AL}{Active Learning}
\newacronym{RL}{RL}{Reinforcement Learning}
\newacronym{MDP}{MDP}{Markov Decision Problem}
\newacronym{NN}{NN}{Neural Network}
\newacronym{KLD}{KLD}{Kullback-Leibler Divergence}
\newacronym{VE}{VE}{Vote Entropy}
\newacronym{UC}{UC}{Uncertainty-Clipping}
\newacronym{Evi}{Evi}{Evidential Neural Networks}
\newacronym{IS}{IS}{Inhibited Softmax}
\newacronym{LS}{LS}{Label Smoothing}
\newacronym{MC}{MC}{Monte-Carlo Dropout}
\newacronym{TeSc}{TeSc}{Temperature Scaling}
\newacronym{TrSc}{TrSc}{TrustScore}

\newacronym{LC}{LC}{Uncertainty Least Confidence}
\newacronym{Rand}{Rand}{Random Sampling}
\newacronym{MM}{MM}{Uncertainty Max-Margin}
\newacronym{Ent}{Ent}{Uncertainty Entropy}
\newacronym{QBC}{QBC}{Query-by-committee}
\newacronym{CAL}{CAL}{Cartography Active Learning}
\newacronym{BALD}{BALD}{Bayesian Active Learning by Disagreement}
\newacronym{core_set}{Core-Set}{Core-Set}
\newacronym{ImitAL}{ImitAL}{Active Learning by Imitation Learning}
\newacronym{ALBL}{ALBL}{Active Learning by Learning}
\newacronym{LAL}{LAL}{Learning Active Learning}
\newacronym{SPAL}{SPAL}{Self-paced Active Learning}
\newacronym{QUIRE}{QUIRE}{QUerying Informative and Representative Examples}
\newacronym{BMDR}{BMDR}{Batch-mode Discriminative and Representative Active Learning}

\glsdisablehyper

\newcommand{\noop}[1]{}

\newcolumntype{C}[1]{>{\centering\arraybackslash}p{#1}}
\newcolumntype{R}[1]{>{\raggedleft\arraybackslash}p{#1}}
\newcolumntype{L}[1]{>{\raggedright\arraybackslash}p{#1}}
\NewDocumentCommand{\rot}{O{30} O{1em} m}{\makebox[#2][l]{\rotatebox{#1}{#3}}}

\begin{document}

\title{Survey of Active Learning Hyperparameters: Insights from a Large-Scale Experimental Grid}
\author{%
\IEEEauthorblockN{Julius Gonsior\IEEEauthorrefmark{1}\orcidlink{0000-0002-5985-4348} \and
Tim Rieß\IEEEauthorrefmark{1} \and
Anja Reusch\IEEEauthorrefmark{2}\orcidlink{0000-0002-2537-9841} \and
Claudio Hartmann\IEEEauthorrefmark{1}\orcidlink{0000-0002-5334-059X} \and
Maik Thiele\IEEEauthorrefmark{3}\orcidlink{0000-0002-1665-977X} \and
Wolfgang Lehner\IEEEauthorrefmark{1}\orcidlink{0000-0001-8107-2775}}
\\
\IEEEauthorblockA{\IEEEauthorrefmark{1}Technische Universität Dresden, \{firstname.lastname\}@tu-dresden.de}
\\
\IEEEauthorblockA{\IEEEauthorrefmark{2}Technion - Israeli Institute of Technology, \{firstname\}@campus.technion.ac.il}
\\
\IEEEauthorblockA{\IEEEauthorrefmark{3}Hochschule für Technik und Wirtschaft Dresden, \{firstname.lastname\}@htw-dresden.de}
}

\markboth{\textsc{arXiv} preprint: Survey of Active Learning Hyperparameters}%
{\textsc{arXiv} preprint: Survey of Active Learning Hyperparameters}

\maketitle

\begin{abstract}
Annotating data is a time-consuming and costly task, but it is inherently required for supervised machine learning. \gls{AL} is an established method that minimizes human labeling effort by iteratively selecting the most informative unlabeled samples for expert annotation, thereby improving the overall classification performance. Even though \gls{AL} has been known for decades~\cite{settlesActiveLearningLiterature2009}, AL is still rarely used in real-world applications. As indicated in the two community web surveys among the NLP community about AL~\cite{tomanekWebSurveyUse2009,rombergHaveLLMsMade2025}, two main reasons continue to hold practitioners back from using AL: first, the complexity of setting AL up, and second, a lack of trust in its effectiveness. We hypothesize that both reasons share the same culprit: the large hyperparameter space of \gls{AL}. This mostly unexplored hyperparameter space often leads to misleading and irreproducible AL experiment results. In this study, we first compiled a large hyperparameter grid of over 4.6 million hyperparameter combinations, second, recorded the performance of all combinations in the so-far biggest conducted AL study, and third, analyzed the impact of each hyperparameter in the experiment results. 
In the end, we give recommendations about the influence of each hyperparameter, demonstrate the surprising influence of the concrete AL strategy implementation, and outline an experimental study design for reproducible AL experiments with minimal computational effort, thus contributing to more reproducible and trustworthy AL research in the future.
\end{abstract}

\section{Introduction}
\IEEEPARstart{T}{he} majority of \gls{ML} projects use the supervised learning paradigm. Annotated or labeled data samples, consisting of input and output pairs, are used in a training phase to optimize the \gls{ML} model for the concrete use case. This fundamentally requires labeled data. One of the biggest cost factors of real-world applications is the expense of annotating the required dataset. For example, the average cost for the common label task of segmenting a single image accurately is 6.40 USD\footnote{According to scale.ai as of December 2021 (as of 2024, the cost is not publicly visible anymore): \scriptsize{\url{https://web.archive.org/web/20210112234705/https://scale.com/pricing}}}. \gls{AL} is a human-in-the-loop technique targeted at reducing the amount of required labeled data, thereby lowering the costs of \gls{ML} projects and therefore enabling more real-world applications to happen.

Despite the increase in popularity of \gls{AL} in the research community~\footnote{According to \url{https://webofscience.com/} there have been more than 5000 publications per year in the \gls{AL} field in the past years}, and the significant cost savings demonstrated~\cite{gonsiorActiveLearningSpreadsheet2020, sassDeepCAVEInteractiveAnalysis2022, nathDiminishingUncertaintyTraining2021, narayananActiveLearningAir2020, loganDECALDEployableClinical2022, kishaanUsingActiveLearning2020, demirUnsupervisedActiveLearning2015}, which could have been achieved if \gls{AL} had been applied correctly, \gls{AL} is still a rare technique and is not widely used by \gls{ML} practicioners~\cite{tomanekWebSurveyUse2009,rombergHaveLLMsMade2025}. Although commercial annotation tools often support \gls{AL}~\footnote{e.g., Labelbox, Prodigy, AWS Comprehend, and AWS SageMaker Ground Truth}, it usually must be enabled manually, often via a plugin. The implemented \gls{AL} strategies are frequently state-of-the-art from the early 90s, ignoring a considerable research body of over 30 years. An anecdotal experience from talking at research conferences to other potential applicants of \gls{AL}: potential practitioners of \gls{AL} often ask either the questions: "But does it \textit{really} work in practice?" or "I would like to apply \gls{AL}, but how should I do it?". This distrust in the ability of \gls{AL} to work is also the main reason for withholding practitioners from using \gls{AL} in the community survey by Tomanek and Olsson~\cite{tomanekWebSurveyUse2009}. Despite the demonstrated success of \gls{AL} in individual use cases, there is no clear consensus on how to set up an \gls{AL} system in the \textit{correct} way. This paper aims to answer the two questions mentioned above.

Our main hypothesis for reasons behind the limited trust and resulting limited adoption of the \gls{AL}'s research results is that empirical evaluation results are often contradictory, not reproducible, and not comparable. This can be illustrated by the unresolved question of "What is the best \gls{AL} strategy?". Several survey papers have attempted to answer this by empirically comparing various \gls{AL} strategies. In the three surveys \cite{yangBenchmarkComparisonActive2018, bahriMarginAllYou2022, schroderRevisitingUncertaintybasedQuery2022} uncertainty-based \gls{AL} strategies were found to perform the best; however, each survey recommends a different uncertainty-based strategy. Meanwhile, in \cite{desreumauxLearningActiveLearning2020, zhanComparativeSurveyBenchmarking2021}, non-uncertainty-based strategies outperform uncertainty-based ones. In contrast, in \cite{schroderRevisitingUncertaintybasedQuery2022}, the Uncertainty Entropy Sampling strategy outperformed random sampling across multiple datasets and scenarios. In contrast, \cite{desreumauxLearningActiveLearning2020,ghoseFragilityActiveLearners2024} found that no strategy consistently performed better than random sampling across all datasets. A good answer to the question about the best strategy is not easy to find. First, a correct answer is very nuanced because the question is too general to have a single, short answer. It depends on many more properties of the actual use case than a simple strategy recommendation. Second, comparing the results of two \gls{AL} experiments is challenging because many hyperparameters significantly influence \gls{AL}. These hyperparameters include the datasets used, evaluated strategies, metrics applied, and more. This paper aims to analyze the influence of the \gls{AL} hyperparameters to ultimately explain the differences in the empirical \gls{AL} evaluations, not to replace them with a simply bigger study. In our evaluation, we will give a possible explanation for the different results in the \gls{AL} literature: that the survey results of two different hyperparameter combinations come to the same conclusion and correlate with each other is only possible by chance. Due to the sheer number of hyperparameter combinations, special care must be taken when defining the hyperparameter grid. We will recommend how to do this at the end of our evaluation.

Our work is not the first to identify that the \gls{AL} research field lacks consensus on the hyperparameters used in empirical evaluations~\cite{jiRandomnessRootAll2023}, a unified benchmark~\cite{margrafALPBenchBenchmarkActive2024}, and agreement on the influence of hyperparameters on \gls{AL} experiments~\cite{evansWhenDoesActive2013}. Despite numerous studies addressing these issues, few have yielded satisfactory conclusions. In Tab.~\ref{tab:related_work} in Sec.~\ref{sec:hyperparameters}, we provide a list of all analyzed hyperparameter values across different studies. Most studies focus on a few hyperparameters, neglecting the interdependency and influence of others. This is the first work to extensively analyze all relevant \gls{AL} hyperparameters in the biggest experimental \gls{AL} survey so far. While previous surveys evaluated up to 140,625 parameter combinations, our final grid search covered over 4.6 million parameters.
Our core contributions are:
\begin{enumerate}
    \item Outlining the complexity of conducting \gls{AL} experiments by compiling all relevant hyperparameters
    \item Conducting an extensive grid search over all identified hyperparameters
    \item Analyzing the resulting large dataset of \gls{AL} experiment results and providing qualitative and empirical insights into how \gls{AL} hyperparameters influence\gls{AL}
    \item Demonstrating the challenges and variety in implementing \gls{AL} strategies
    \item Giving recommendations, which \gls{AL} hyperparameters should be used in \gls{AL} experiments for trustworthy and reproducible results
\end{enumerate}

Our paper is organized as follows: first, we provide a basic introduction to \gls{AL} (Sec.~\ref{sec:al_basics}). We then give a more profound overview of the possible hyperparameters of a typical \gls{AL} setting in Sec.~\ref{sec:hyperparameters}. Afterwards, we explain how we conducted our experiments in Sec.~\ref{sec:implementation}, followed by the analysis in Sec.~\ref{sec:evaluation}.
\section{Active learning experiment hyperparameters and fundamentels}
\label{sec:al_basics}
This section introduces the basic framework of the \acrfull{AL} cycle in Sec.~\ref{sec:al_cycle} and outlines the hyperparameters of \gls{AL} experiments in Sec.~\ref{sec:hyperparameters}.
\subsection{Active learning cycle}
\label{sec:al_cycle}
Supervised learning methods depend fundamentally on annotated datasets. \gls{AL} is an established method that minimizes human labeling effort by iteratively selecting the most informative unlabeled samples for expert annotation, thereby improving the overall classification performance~\cite{settlesActiveLearningLiterature2009}. The goal of \gls{AL} is to train a classification model $\learner$ that maps samples $x \in \mathbb{X}$ to labels $y \in \mathbb{Y}$, with labels $\mathbb{Y}$ provided by an \textit{oracle}, usually human experts.
\begin{figure}[!htp]
    \centering
    \includegraphics[width=3in]{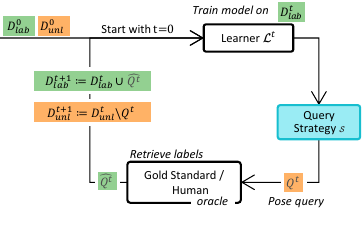}
    \caption{Standard Active Learning Cycle}
    \label{fig:al_cycle}
\end{figure}
Fig~\ref{fig:al_cycle} illustrates a typical pool-based \gls{AL} cycle, the cycle iteration is indicated by $t$: Starting with a small initial labeled dataset $\labeledSet^{0} = \{(x_i,y_i)\}_{i=0}^\amountOfSamples$ of $n$ samples $x_i \in \mathbb{X}$ and corresponding labels $\singleLabel_i \in \mathbb{Y}$, and a large pool of unlabeled data $\unlabeledSet^0 = \{x_i\}, x_i \not\in \labeledSet^0$, an ML model, called the \textit{learner} model $\learner\colon \mathbb{X} \mapsto \mathbb{Y}$, is trained on the labeled data. A \emph{query strategy} (blue box) $\ALStrategy\colon \unlabeledSet \longrightarrow \unlabeledQuery$ selects a batch of $\batchSize$ unlabeled samples $\unlabeledQuery^t \subseteq \unlabeledSet^t$ for labeling by the oracle (lower box), and these newly labeled samples $\widehat{\unlabeledQuery^t}$ are added to the labeled dataset to form the newly labeled dataset for the next \gls{AL} cycle $\labeledSet^{t+1}$, as well as removed from the next unlabeled set $\unlabeledSet^{t+1}$. This cycle repeats $\AmountOfTrainingALCycles$ times until a predefined stopping criterion is reached. The results $\ALExperimentResult$ of an \gls{AL} process can be recorded as a time series, either containing a metric $\ALMetric$ measured at each time step $\ALExperimentResult(\ALMetric)$, or the set of queried samples for labeling $\ALExperimentResult(\unlabeledQuery)$:
\begin{align*}
    \ALExperimentResult(\ALMetric) &= [\ALMetric^0, \dots, \ALMetric^\AmountOfTrainingALCycles]\\
    \ALExperimentResult(\unlabeledQuery) &= [\unlabeledQuery^0, \dots, \unlabeledQuery^\AmountOfTrainingALCycles]
\end{align*}
\subsection{Hyperparameters of \acrlong{AL} experiments}
\label{sec:hyperparameters}
In this paper we define a single \gls{AL} experiment $\ALExperiment=(\ALStrategy, \ALDataset, \ALTrainTestSplit, \ALStartSet, \ALMetric, \batchSize, \AmountOfTrainingALCycles, \learner)$ as a combination of hyperparameters, influencing the simulation of one \gls{AL} strategy $\ALStrategy$ on a prelabeled dataset $\ALDataset$. Given a pre-labeled dataset $\ALDataset = \{(x_i,y_i)\}_{i=0}^\amountOfSamples$, consisting of $\amountOfSamples$-data samples $\singleSample$ and corresponding labels $\singleLabel$, the data set is first split into a disjoint training set $\trainSet$ and test set $\testSet$ using the train-test-split $\ALTrainTestSplit(\ALDataset) = (\trainSet,\testSet)$. The train set is further divided into an initial labeled start set $\labeledSet$, and an unlabeled set $\unlabeledSet$ using the initial start set function $\ALStartSet(\trainSet) = (\labeledSet, \unlabeledSet)$. During the simulation, one or more metrics $\ALMetric$ typically measure how much human effort \textit{could have been} saved using \gls{AL} from the beginning. This is done by utilizing the test set $\testSet$. The batch size $\batchSize$ describes the \gls{AL} batch size, while $\AmountOfTrainingALCycles$ indicates how many times the \gls{AL} cycle is repeated. The learner model  $\learner$ is the \gls{ML} model used.

A series of multiple experiments is denoted as $\ALExperimentSet = \{\ALExperiment_0, \cdots, \ALExperiment_i\}$. An experimental evaluation grid, consisting of multiple values for each hyperparameter, can be written as a cross product of the following hyperparameter sets: \\$\ALStrategySet \times \ALDatasetSet \times \ALTrainTestSplitSet \times \ALStartSetSet  \times \batchSizeSet \times \learnerSet$.
\begin{align*}
    \ALDatasetSet&=\{\ALDataset_0, \cdots, \ALDataset_i\} \text{, dataset}\\
    \ALTrainTestSplitSet&=\{\ALTrainTestSplit_0, \cdots, \ALTrainTestSplit_j\} \text{, train-test-split}\\
    \ALStartSetSet&=\{\ALStartSet_0, \cdots \ALStartSet_k\} \text{, initial start set}\\
    \ALStrategySet&=\{\ALStrategy_0, \cdots, \ALStrategy_l\} \text{, \gls{AL} strategy}\\
    \batchSizeSet&=\{\batchSize_0, \cdots, \batchSize_m\} \text{, batch size}\\
    \learnerSet&=\{\learner_0, \cdots, \learner_n\} \text{, \gls{ML} learner model}\\
    \AmountOfTrainingALCycles&=\{\AmountOfTrainingALCycles_0, \cdots, \AmountOfTrainingALCycles_o\} \text{, amount of training cycles}\\
    \ALMetricSet&=\{\ALMetric_0, \cdots, \ALMetric_p\} \text{, metric}
\end{align*}

The hyperparameter sets of the metrics $\ALMetricSet$ and the amounts of \gls{AL} cycles $\AmountOfTrainingALCyclesSet$ of an \gls{AL} experiment $\ALExperiment$ are not part of the experimental grid, as metrics can be calculated simultaneously for the same experiment, and the maximum number of \gls{AL} cycles can be evaluated as an ablation study post hoc. Both parameters do not need a re-run of the same experiment. 
The results $\ALExperimentResult$ of a single \gls{AL} experiment $\ALExperiment$ are denoted as a time series over the \gls{AL} cycles from $0$ to $\AmountOfTrainingALCycles$, each time either containing the results for an evaluation metric $\ALMetric$: $\ALExperimentResult(\ALMetric) = [\ALMetric^0, …, \ALMetric^\AmountOfTrainingALCycles]$, or the sets of selected batches for labeling $\unlabeledQuery$: $\ALExperimentResult(\unlabeledQuery) = [\unlabeledQuery^0, …, \unlabeledQuery^\AmountOfTrainingALCycles]$. The results of a series of experiments $\ALExperimentSet$ are denoted as $\ALExperimentResultsSet=\{\ALExperimentResult_\ALExperiment | \ALExperiment \in \ALExperimentSet\})$. Since not all grid combinations of the grid are evaluated~\footnote{e.g. for large datasets the batch size is often higher compared to smaller datasets}, the result set $\ALExperimentResultsSet$ does not always contain the results for the complete possible grid of $\ALDatasetSet \times \ALTrainTestSplitSet \times \ALStartSetSet \times \ALStrategySet \times \batchSizeSet \times \learnerSet$.

Our list of hyperparameters is not exhaustive. We have focused on the most common parameters in most evaluation sections of \gls{AL} research papers. Additional hyperparameters could include, e.g., the oracle (one oracle vs. many oracle, handling of oracle conflicts, label noise,… ) or the stopping criteria used.
\subsection{Overview of evaluated parameter ranges in related work}
There has been an ongoing trend in the past decade, with many papers analyzing the influence of \gls{AL} hyperparameters, primarily by conducting large-scale empirical analyses of  \gls{AL} strategies. We have gathered an overview of the latest research in Tab.~\ref{tab:related_work} by summarizing, to our best effort, all evaluated hyperparameters in the respective papers. Some papers resulted in large empirical evaluation grids, while others focused on introducing new hyperparameter dimensions.
We estimate the size of the experimental hyperparameter grid in the column $|\ALExperimentResultsSet|$. As the compared concrete parameters vary significantly among the papers, this number merely estimates the number of compared hyperparameter combinations. It should not be used to compare computational or research efforts directly. 

Surprisingly, despite the large body of research on this topic and the high expectations at the beginning of each endeavor, most studies could only partially answer their research questions, leaving many unresolved issues about \gls{AL}. We hypothesize that focusing on single hyperparameters reveals little, as the most significant effort occurs between them, necessitating a large hyperparameter experimental grid.

\subsubsection{Hyperparameter comparison works}
We will start by giving an overview of the related works about analysis of the hyperparameters of \gls{AL} and recap their main findings related to our research goal. All share the same motivation to investigate the hyperparameter space of \gls{AL}, with each paper having a different focus, often on individual hyperparameters. Due to the emphasis on a few hyperparameters, the results are always influenced by the arbitrary choice of the unexplored hyperparameters.

\textbf{Importance of hyperparameters:}
\cite{evansWhenDoesActive2013} were among the first to conclude that \gls{AL} experiments are very complex, with many hyperparameters influencing them. Random sampling beat most \gls{AL} strategies in 89\% of their experiments. On the same note, the paper~\cite{kottkeChallengesReliableRealistic2017} outlines the evaluation challenges of \gls{AL}, focusing on reliable, realistic, and comparable results.

\textbf{Learner model:}
\cite{pereira-santosEmpiricalInvestigationActive2019} found that the chosen learner model significantly influences experiment outcomes. However, they conclude that "the uncovering of the relationships between dataset characteristics, learning algorithms, and active learning methods is still needed." 

\textbf{Metric:}
The focus of \cite{ramirez-loaizaActiveLearningEmpirical2017} was on comparing performance metrics. They found that most strategies excel in individual performance metrics but perform poorly in others. 
\cite{abrahamRebuildingTrustActive2020} introduced new metrics for monitoring the performance of ongoing \gls{AL} processes.

\textbf{Dataset:}
\cite{yangBenchmarkComparisonActive2018} were the first to use a larger number of datasets and concluded that \gls{AL} strategies still do not consistently outperform random sampling.

\textbf{\gls{AL} strategy:}
The large survey~\cite{zhanComparativeSurveyBenchmarking2021} compared almost all existing \gls{AL} strategies but provided limited insight into why specific strategies perform better on some datasets than others. \cite{luReBenchmarkingPoolBasedActive2023} repeated the benchmark of \cite{zhanComparativeSurveyBenchmarking2021}, and found surprisingly significant discrepancies between their results and the original findings. They noted that "more than half of the examined strategies do not exhibit significant advantages over the [random] baseline." \cite{bahriMarginAllYou2022} recommended strongly the max-margin-based uncertainty-sampling \gls{AL} strategy, contrary to findings from other papers. 

\textbf{Start set selection strategy:}
\cite{chandraInitialPoolsDeep2021} compared strategies for selecting the initial labeled start sets and found that random selection performs as well as computationally more heavy and conceptionally complex strategy. 

\textbf{\gls{AL} for outlier detection:}
\cite{trittenbachOverviewBenchmarkActive2021} focused on comparing \gls{AL} for \textit{outlier detection} and \textit{one class classifiers}, using an extensive evaluation hyperparameter combinations. 

\textbf{Label noise:}
\cite{abrahamSampleNoiseImpact2021} explored the often overlooked hyperparameter of label noise introduced by the oracle. Although noise significantly affects performance, they could not identify any strategy as being more noise-robust. Our interpretation of this parameter and their results is that noise impacts all \gls{AL} strategies and could be mitigated by labeling guidelines or multiple annotators. 

\textbf{Stopping criteria:}
\cite{pullar-streckerHittingTargetStopping2022} performed the most extensive study on different stopping criteria of \gls{AL} to stop the process earlier than the given budget. As their extensive qualitative study seems to answer all questions about this parameter and also seems not to have any interdependencies with other parameters, we decided not to include this parameter in our study. 

\textbf{Deep learning:}
Newer research, such as~\cite{zhanComparativeSurveyDeep2022, jiRandomnessRootAll2023, margrafALPBenchBenchmarkActive2024}, focuses on deep-learning based \gls{ML} learner models, often resulting in new \gls{AL} strategies, especially in how uncertainty is calculated for deep neural networks.

\begin{table*}[]
\addtolength{\tabcolsep}{-5pt}
\begin{minipage}{\textwidth}
\begin{tabular}{L{4cm}R{0.5cm}R{2.5cm}R{1.5cm}R{1.2cm}R{1.5cm}R{1.2cm}R{2cm}R{1.5cm}R{1.4cm}}
\toprule
 \multicolumn{2}{r}{\#Datasets $|\ALDatasetSet|$}&
  \#Start Sets $|\ALStartSetSet|$ $\times$ Train-Test Split $|\ALTrainTestSplitSet|$&
  $|\ALStartSetSet \times \ALTrainTestSplitSet|$&
  \#ML Model $|\learnerSet|$&
  \#Query Strategy $|\ALStrategySet|$&
  \#Batch Size $|\batchSizeSet|$&
  \#Metrics $|\ALMetricSet|$&
  \#$|\ALExperimentResultsSet|$ \\ 
\midrule

Schein et. al. 2007~\cite{scheinActiveLearningLogistic2007} &
  10 &
  10 $\times$ 10 &
  \textit{100} &
  1 &
  7 &
  1 &
  6 &
  7,000 \\
Evans et. al. 2013~\cite{evansWhenDoesActive2013} &
  4 &
  4 $\times$ 1 &
  4 &
  4 &
  2 &
  1 &
  3 &
  128 \\
Pereira-Santos et. al. 2017~\cite{pereira-santosEmpiricalInvestigationActive2019} &
  75 &
  5 $\times$ 5 &
  25 &
  5 &
  15 &
  1 &
  3 &
  140,625 \\
Ramirez-Loaiza et. al. 2017~\cite{ramirez-loaizaActiveLearningEmpirical2017} &
  10 &
  10 $\times$ 5 &
  50 &
  2 &
  3 &
  1 &
  5 &
  3,000 \\
Kottke et. al. 2017~\cite{kottkeChallengesReliableRealistic2017} &
  1 &
  10 $\times$ 10 &
  \textit{100} &
  1 &
  4 &
  1 &
  1 &
  400 \\
Yang et. al. 2018~\cite{yangBenchmarkComparisonActive2018} &
  47 &
  20/1,000$\times$ 1~\footnote{20 for 44 real-world datasets, 1,000 repetitions for 3 synthetic datasets} &
  20/\textbf{1,000} &
  1 &
  9 &
  1 &
  4 &
  34,920 \\
Abraham et. al. 2020~\cite{abrahamRebuildingTrustActive2020} &
  9 &
  10 $\times$ 5 &
  50 &
  1 &
  6 &
  2 &
  1 &
  54,000 \\
Zhan et. al. 2021~\cite{zhanComparativeSurveyBenchmarking2021} &
  35 &
  10/100 $\times$1~\footnote{For 27 datasets with less than 2,000 samples the start set was set to 10, for the other 8 to 100} &
  10/\textit{100} &
  1 &
  \textit{19} &
  \textit{4} &
  2 &
  81,320 \\
Chandra et. al. 2021~\cite{chandraInitialPoolsDeep2021} &
  4 &
  5 $\times$1 &
  5 &
  1 &
  7 &
  1 &
  3 &
  140 \\
Trittenbach et. al. 2021~\cite{trittenbachOverviewBenchmarkActive2021} &
  60~\footnote{Based on 20 real datasets, which are each augmented to include random outliers three times} &
  3 $\times$ 4 / 1 $\times$ 5~\footnote{Two scenarios with different splitting strategies were evaluated} &
  12 / 5 &
  \textbf{10} / 4~\footnote{In the first scenario 5 base learners with 2 kernel parameter nationalizations were used, in the other only two base learners}& 
  10 &
  1 &
  \textit{8} &
  84,000 \\
Abraham et. al. 2021~\cite{abrahamSampleNoiseImpact2021} &
  10 &
  10 $\times$5 &
  50 &
  2 &
  4 &
  1 &
  2 &
  4,000 \\
Pullar-Strecker et. al. 2022~\cite{pullar-streckerHittingTargetStopping2022} &
  9 &
  30 $\times$1 &
  30 &
  3 &
  1 &
  1 &
  1 &
  810 \\
Bahri et. al. 2022~\cite{bahriMarginAllYou2022} &
  69 &
  20 $\times$1 &
  20 &
  3 &
  16 &
  1 &
  2 &
  66,240 \\
Zhan et. al. 2022~\cite{zhanComparativeSurveyDeep2022} &
  10 &
  1 $\times$ 3 &
  3 &
  1 &
  \textit{19} &
  \textit{4} &
  2 &
  2 280 \\
Lu et. al. 2023~\cite{luReBenchmarkingPoolBasedActive2023} &
  26 &
  10/100 $\times$1~\footnote{For 20 datasets with less than 2,000 samples the start set was set to 10, for the other 6 to 100} &
  10/\textit{100} &
  1 &
  17 &
  1 &
  2 &
  13,600 \\
Ji et. al. 2023~\cite{jiRandomnessRootAll2023} &
  2 &
  10 $\times$1 &
  10 &
  1 &
  7 &
  3 &
  2 &
  420 \\
Margraf et. al. 2024~\cite{margrafALPBenchBenchmarkActive2024} &
  \textit{86} &
  2 $\times$10 &
  20 &
  \textit{8} &
  9 &
  1 &
  3 &
  123,840 \\
Ours &
  \textbf{92} &
  20 $\times$5 &
  \textit{100} &
  3 &
  \textbf{28} &
  \textbf{6} &
  \textbf{49} &
  4,636,800 \\
\bottomrule
\end{tabular}
\caption{Overview of evaluated Hyperparameter combinations in \gls{AL} survey papers. Note that the total number of evaluated hyperparameter combinations $|\ALExperimentResultsSet|$ is in most of the papers probably smaller than reported here as the complete experimental grid has been completely computed, and some parameters have only been evaluated on a smaller subset. Bold numbers indicate the highest number of compared parameters, and italic numbers are the second-highest compared parameters.}
\label{tab:related_work}
\end{minipage}
\end{table*}
\subsubsection{Common hyperparameter ranges}
After listing the related works on \gls{AL} hyperparameter investigations, we are compiling in this section an overview of the common hyperparameter values used in \gls{AL} literature to use as a basis for the decision of our hyperparameter grid.

Common dataset $\ALDatasetSet$ choices include the UCI Machine Learning Repository~\cite{kellyUCIMachineLearning2017}, the Open ML Benchmarking Suite CC18~\cite{bischlOpenMLBenchmarkingSuites2021}, synthetic datasets with a specific property like an XOR-like dataset~\cite{konyushkovaLearningActiveLearning2017, gonsiorComparingImprovingActive2024} or domain-specific datasets like ImageNet~\cite{dengImageNetLargescaleHierarchical2009}. The evaluated parameters values for the set of \gls{ML} learner models $\learnerSet$ often include general-purpose \gls{ML} models such as multi-layer perceptrons, random forest classifiers, support vector machines, logistic regression, or domain-specific neural network architectures like transformer-models~\cite{schroderRevisitingUncertaintybasedQuery2022}. 

The set of query strategies $\ALStrategySet$ typically includes random sampling as a baseline and a mixture of informativeness-based and representativeness-based query strategies. A simple uncertainty-based strategy like Least Confidence~\cite{settlesActiveLearningLiterature2009} is often used. The batch size parameter is usually not rigorously evaluated, with researchers guessing real-world values for respective datasets, ranging from 1~\cite{scheinActiveLearningLogistic2007} to values of up to 10,000~\cite{zhanComparativeSurveyDeep2022}. The train-test-split function set $\ALTrainTestSplitSet$ varies across papers, with some opting for n-times k-fold cross-validation, while others use fixed splits with more different starting sets. The hyperparameter $\AmountOfTrainingALCycles$ defining the end of the \gls{AL} cycle is rarely evaluated. Some papers use a fixed number of cycles, ranging from 20~\cite{scheinActiveLearningLogistic2007} to the dataset's length~\cite{zhanComparativeSurveyDeep2022}. Few employ stopping criteria~\cite{pullar-streckerHittingTargetStopping2022}.
\subsubsection{Special case of \gls{AL} evaluation metric}
\begin{figure*}[!htp]
\centering
\subfloat[Overfull plot using 28 strategies (and the statistical variation over multiple train-test-splits or multiple start sets)]{\includegraphics[width=2.5in]{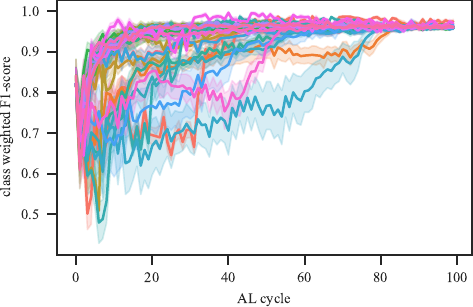}%
\label{fig:single_learning_curve_real}}
\hfil
\subfloat[Debatable interpretations of which strategy is the better one here]{\includegraphics[width=2.5in]{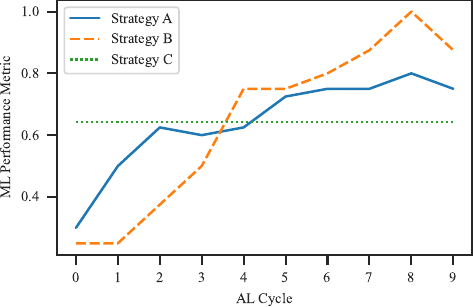}%
\label{fig:single_learning_curve_fake}}
\caption{Showcasing the limits of learning curve plots}
\label{fig:single_learning_curves}
\end{figure*}
The evaluation metric used in \gls{AL} experiments varies significantly across studies, which deserves special attention. The main goal of a metric is to either quantify the performance of the \gls{AL} strategy, reflecting the saved human effort, or to gain insights into how the strategy functions. The challenge lies in maximizing the human effort saving with minimal performance loss for the learner model due to less available labeled data~\cite{kottkeChallengesReliableRealistic2017, settlesActiveLearningLiterature2009, kremplProbabilisticActiveLearning2014}. 

Most researchers use standard \gls{ML} metrics such as accuracy, precision, or the F1-score to measure the learner \gls{ML} model's performance on the currently available labeled data. As an \gls{AL} process can be seen as a graph, the most common way to compare multiple experiments are so-called "learning curve"-plots. Fig.~\ref{fig:single_learning_curves} shows exemplary learning curves, highlighting common issues in using learning curves for evaluation. In Fig.~\ref{fig:single_learning_curve_real}, the plot becomes cluttered with 28 strategies, while Fig.~\ref{fig:single_learning_curve_fake} illustrates the ambiguity of determining which strategy performs best. Strategy A requires fewer labeled samples at the beginning to reach a high metric but stagnates early on. Strategy B requires more labeled samples in the beginning to achieve the same metric value but reaches higher values in the end with a growing set of labeled samples. Strategy C, in contrast, has a constant value all the time. It is debatable if the poor performance of Strategy B in the beginning in contrast to Strategy A outweighs the better performance in the last \gls{AL} cycles. The main reason behind the difficulty of interpreting the learning curve plots lies in the problem of combining the two objectives of \gls{AL}, saving as much human effort as possible, but at the same time training a high-performing learner model~\cite{kremplProbabilisticActiveLearning2014, kottkeChallengesReliableRealistic2017}.

An abundance of methods exists to \textit{aggregate} learning curves into a single quantifiable value, referred to as \textit{aggregation}-metrics. An easy solution is to focus only on the final \gls{AL} iteration's metric value, given a fixed budget~\cite{schroderRevisitingUncertaintybasedQuery2022}. However, this  ignores the progress in fluctuating \gls{AL} learning curves. A more popular approach is to take the arithmetic mean over all \gls{AL} cycles~\cite{trittenbachOverviewBenchmarkActive2021, gonsiorImitALLearnedActive2022, schroderRevisitingUncertaintybasedQuery2022}. This is similar to the area-under-curve metrics, which interpolate the integral using the trapezoidal rule. As already stated in other works such as~\cite{trittenbachOverviewBenchmarkActive2021, gonsiorComparingImprovingActive2024}, approximating the integral using the trapezoidal rule is not necessary (and even potentially misleading), as one can directly calculate the integral of the discrete function using the arithmetic mean. Putting further emphasis on the discovery of Evans et al. ~\cite{evansWhenDoesActive2013} that most of the \gls{AL} gain occurs in the early \gls{AL} cycles, some work further partitions the learning curve into multiple areas~\cite{trittenbachOverviewBenchmarkActive2021, kottkeChallengesReliableRealistic2017}. The first \gls{AL} cycles are called \textit{ramp-up}- or \textit{early}-phase, the last ones the \textit{plateau}- or \textit{saturation}-phase. Unfortunately, most works use a fixed threshold in their evaluation to differentiate between these phases, which is impractical since this is a highly dataset-dependent threshold.

In our evaluation, we analyze the influence of different ranges on the outcome of \gls{AL} evaluations, and inspired by~\cite{reitmaierLetUsKnow2013, kottkeChallengesReliableRealistic2017}, we propose a dynamic approach to distinguish the ramp-up- from the plateau-phase during \gls{AL} experiments using the performance of the random strategy as a baseline.

As standard \gls{ML} metrics such as accuracy or the F1-score are only partially comparable between two datasets (e.g., a 5\% gain for one dataset is not the same as a 20\% gain for another), some researchers prefer to use \textit{ranks} instead of absolute ML metric values~\cite{kottkeChallengesReliableRealistic2017, gonsiorImitALLearnedActive2022}. The ranks can be presented in a tabular performance report or as part of a learning curve.

Many studies also incorporate statistical testing in their evaluation to demonstrate that a potential strategy's superiority is just due to random chance. Typical tests include the t-test~\cite{caiActiveLearningSupport2014, huangNovelUncertaintySampling2016, yangActivelyInitializeActive2022, jiRandomnessRootAll2023} and its variants, such as Welch's t-test~\cite{bahriMarginAllYou2022}, the more popular non-parametric Wilcoxon signed rank test~\cite{wilcoxonIndividualComparisonsRanking1945,caiActiveLearningSupport2014, huangActiveLearningQuerying2010, kremplOptimisedProbabilisticActive2015, sonActiveLearningUsing2016}, or the Friedman test~\cite{demsarStatisticalComparisonsClassifiers2006, abrahamRebuildingTrustActive2020}, which takes differences between datasets into account. Results are often presented as win-/tie-/loss-matrices~\cite{kremplOptimisedProbabilisticActive2015, yangBenchmarkComparisonActive2018, desreumauxLearningActiveLearning2020, bahriMarginAllYou2022}. Common significance levels are 90\%~\cite{yangBenchmarkComparisonActive2018, jiRandomnessRootAll2023}, 95\%~\cite{abrahamRebuildingTrustActive2020, yangActivelyInitializeActive2022}, and 99\%~\cite{desreumauxLearningActiveLearning2020, bahriMarginAllYou2022}.

Given that \gls{AL} strategies can exhibit quadratic runtime complxeity~\cite{luReBenchmarkingPoolBasedActive2023, gonsiorImitALLearnedActive2022}, runtime is another favorable metric for display~\cite{gonsiorImitALLearnedActive2022, schroderRevisitingUncertaintybasedQuery2022}.
\subsection{Parameter grid used in our experiments}
\label{sec:our_param_grid}
\begin{table}[]
\centering
\begin{minipage}{.5\textwidth}
\begin{tabular}{L{2.2cm}L{5.5cm}}
\toprule
Parameter name     & Parameter Values                        \\ \midrule
Dataset~$\ALDatasetSet$           & 92 Datasets in total from UCI~\cite{kellyUCIMachineLearning2017}, Kaggle~\cite{KaggleYourMachine} and OpenML~\cite{bischlOpenMLBenchmarkingSuites2021}, ranging from 100 to 20,000 samples, from 2 to 31 classes, and from 2 to 1 776 feature dimensions~\footnote{The complete list of used datasets can be found in the source code repository for this paper under \url{https://github.com/jgonsior/olympic-games-of-active-learning/tree/main/dataset_list.txt}}\\
Train-Test-Splits~$\ALTrainTestSplitSet$ & 5 random splits per dataset                                \\
Start Point Sets~$\ALStartSetSet$ & 20 random start sets per dataset train test split, each with one example per classification class \\
AL Strategies~$\ALStrategySet$    & 28 different implementations taken from combining ALiPy~\cite{tangALiPyActiveLearning2019}, libact~\cite{yangLibactPoolbasedActive2017}, Google Playground~\cite{yangGoogleActivelearning2024}, scikit-activeml~\cite{kottkeScikitactivemlLibraryToolbox2021}, Small-Text~\cite{schroderSmallTextActiveLearning2023}, see Table~\ref{tab:al_strats} for further details  \\
Batch Sizes~$\batchSizeSet$  & 1, 5, 10, 20, 50, 100                       \\
Learner Models~$\learnerSet$   & Neural Network (MLP), Support-Vector-Machines (SVM) with RBF Kernel, Random Forest Classifier (RF)                 \\
Metrics~$\ALMetricSet$          & \textit{Basic \gls{ML} metrics:} accuracy, class-weighted F1-score, class-weighted precision, class-weighted recall, macro F1-score, macro precision, macro recall, runtime, queried data samples \newline \textit{Aggregation metrics, applicable per basic \gls{ML} metric:} full mean, first 5, last 5, last value, ramp-up, plateau                     \\
\bottomrule
\end{tabular}
\end{minipage}
\caption{Hyperparameter Grid used by us}
\label{tab:my_param_grid}
\end{table}

\begin{table}[]
\addtolength{\tabcolsep}{-3.5pt}
\begin{tabular}{L{3cm}lll}
\toprule
Strategy Name     & Framework                                                     & Abbreviation & Source                                                             \\ \midrule
ConstrastiveAL    & Small-Text~\cite{schroderSmallTextActiveLearning2023}         & CAL          & \cite{margatinaActiveLearningAcquiring2021}                        \\
Cost Embedding    & scikit-activeml~\cite{kottkeScikitactivemlLibraryToolbox2021} & CE           & \cite{huangNovelUncertaintySampling2016}                           \\
Coreset Greedy    & ALiPy~\cite{tangALiPyActiveLearning2019}                      & CORE         & \cite{senerActiveLearningConvolutional2018}                        \\
Coreset Greedy    & Small-Text~\cite{schroderSmallTextActiveLearning2023}         & CORE         & \cite{senerActiveLearningConvolutional2018}                        \\
Density Weighted  & ALiPy~\cite{tangALiPyActiveLearning2019}                      & DWUS         & \cite{donmezDualStrategyActive2007, nguyenActiveLearningUsing2004}                            \\
Density Weighted  & libact~\cite{yangLibactPoolbasedActive2017}                   & DWUS         & \cite{donmezDualStrategyActive2007, nguyenActiveLearningUsing2004}                            \\
Embedding K-Means & Small-Text~\cite{schroderSmallTextActiveLearning2023}         & EKM          & \cite{yuanColdstartActiveLearning2020}                             \\
Graph Density     & ALiPy~\cite{tangALiPyActiveLearning2019}                      & GD           & \cite{ebertRALFReinforcedActive2012}                               \\
Optimal Greedy 10 & Done by us                                                    & OG10         & \cite{gonsiorImitALLearnedActive2022, liuLearningHowActively2018} \\
Optimal Greedy 20 & Done by us                                                    & OG20         & \cite{gonsiorImitALLearnedActive2022, liuLearningHowActively2018} \\
Query-by-Committee Kullback-Leibler Divergence &
  scikit-activeml~\cite{kottkeScikitactivemlLibraryToolbox2021} &
  QBC KL &
  \cite{seungQueryCommittee1992, abeQueryLearningStrategies1998, burbidgeActiveLearningRegression2007} \\
Query-by-Committee Vote Entropy &
  scikit-activeml~\cite{kottkeScikitactivemlLibraryToolbox2021} &
  QBC VE &
  \cite{seungQueryCommittee1992, abeQueryLearningStrategies1998, burbidgeActiveLearningRegression2007} \\
QUIRE             & libact~\cite{yangLibactPoolbasedActive2017}                   & QUIRE        & \cite{huangActiveLearningQuerying2010}                             \\
QUIRE             & scikit-activeml~\cite{kottkeScikitactivemlLibraryToolbox2021} & QUIRE        & \cite{huangActiveLearningQuerying2010}                             \\
Random            & ALiPy~\cite{tangALiPyActiveLearning2019}                      & RAND         & -                                                                  \\
Random            & Small-Text~\cite{schroderSmallTextActiveLearning2023}         & RAND         &                                                                    \\
Uncertainty Entropy           & ALiPy~\cite{tangALiPyActiveLearning2019}                      & Unc\_ENT          & \cite{holubEntropybasedActiveLearning2008}                         \\
Uncertainty Entropy           & libact~\cite{yangLibactPoolbasedActive2017}                   & Unc\_ENT          & \cite{holubEntropybasedActiveLearning2008}                         \\
Uncertainty Entropy           & Small-Text~\cite{schroderSmallTextActiveLearning2023}         & Unc\_ENT          & \cite{holubEntropybasedActiveLearning2008}                         \\
Uncertainty Entropy           & scikit-activeml~\cite{kottkeScikitactivemlLibraryToolbox2021} & Unc\_ENT          & \cite{holubEntropybasedActiveLearning2008}                         \\

Uncertainty Least Confidence  & ALiPy~\cite{tangALiPyActiveLearning2019}                      & Unc\_LC           & \cite{lewisSequentialAlgorithmTraining1994}                        \\
Uncertainty Least Confidence  & libact~\cite{yangLibactPoolbasedActive2017}                   & Unc\_LC           & \cite{lewisSequentialAlgorithmTraining1994}                        \\
Uncertainty Least Confidence  & Small-Text~\cite{schroderSmallTextActiveLearning2023}         & Unc\_LC           & \cite{lewisSequentialAlgorithmTraining1994}                        \\
Uncertainty Least Confidence &
  scikit-activeml~\cite{kottkeScikitactivemlLibraryToolbox2021} &
  Unc\_LC &
  \cite{lewisSequentialAlgorithmTraining1994} \\
Uncertainty Max-Margin        & ALiPy~\cite{tangALiPyActiveLearning2019}                      & Unc\_MM           & \cite{schefferMiningWebActive2001}                                 \\
Uncertainty Max-Margin        & scikit-activeml~\cite{kottkeScikitactivemlLibraryToolbox2021} & Unc\_MM           & \cite{schefferMiningWebActive2001}                                 \\
Uncertainty Smallest-Margin     & libact~\cite{yangLibactPoolbasedActive2017}                   & Unc\_SM           & \cite{tongluoActiveLearningRecognize2004}                          \\
Uncertainty Smallest-Margin     & Small-Text~\cite{schroderSmallTextActiveLearning2023}         & Unc\_SM           & \cite{tongluoActiveLearningRecognize2004}                          \\ \bottomrule
\end{tabular}
\caption{Abbreviations and sources of our  used \gls{AL} strategies}
\label{tab:al_strats}
\end{table}
After extensive research on the previously used hyperparameter ranges, a common trend is that most works have neglected at least a single hyperparameter and usually only focused on a few hyperparameters. This means that all results are based on a single choice for the neglected hyperparameters and are, therefore, ignorant of the influence of this hyperparameter. We aim to solve this by including as many hyperparameter values in our grid as possible to gain deeper insights into their interdependencies. Our used hyperparameters are listed in Table~\ref{tab:my_param_grid} and Table~\ref{tab:al_strats}, totaling in $\ALExperimentResultsSet=4,636,800$ hyperparameter combinations. For datasets, we included as many classification datasets as we could find, ranging from easy to hard datasets, from binary to up to 31-multi-class datasets, from small feature vector spaces with two dimensions to very large dimensions (1,776), and small datasets (100 samples) to large datasets (20,000 samples). Similar to other works, we used five random train-test splits to run each dataset five times. Combined with 20 different start point sets per train-test split, each dataset was repeatedly used 100 times.

We combined the \gls{AL} frameworks ALiPy~\cite{tangALiPyActiveLearning2019}, libact~\cite{yangLibactPoolbasedActive2017}, Google Playground~\cite{yangGoogleActivelearning2024}, scikit-activeml~\cite{kottkeScikitactivemlLibraryToolbox2021}, and Small-Text~\cite{schroderSmallTextActiveLearning2023} to leverage as many publicly implemented \gls{AL} strategies as possible. Unfortunately, some strategies resulted in errors across various datasets, were designed only for binary datasets, or had excessively high runtime complexity. As a result, we ultimately had access to 28 strategies listed in Table~\ref{tab:al_strats}. Although some strategies were implemented multiple times in different frameworks, we treated them as distinct strategies for comparison. Additionally, they can be used as a safeguard to check if our expectations of "different implementations, but the same strategy should equal the same experiment outcome" hold. The random strategy was included as a baseline, and the Optimal Greedy strategy served as an upper bound on performance. The Optimal Greedy strategy had access to a fully labeled set in our simulation and selected the samples that would gain the greatest accuracy in the next \gls{AL} iteration. This \textit{peeking} into the future is only possible in the simulations where the true labels are known. We implemented this strategy in two versions: selecting the best batch from 10 random batches and using 20 random batches.

Regarding batch sizes, we used a variety of values to cover different \gls{AL} scenarios, from small to large datasets. We deliberately did not include dataset or domain-specific learner models, opting for simple yet well-established base learners that have been shown to work in many scenarios and have fast runtime performance. Additionally, calculating uncertainty for models like Support-Vector-Machines and Random Forest Classifier is sound, compared to the various ways of computing uncertainty for Neural Networks~\cite{karamchetiMindYourOutliers2021, gonsiorComparingImprovingActive2024}.

\begin{figure*}[!htp]
\centering
\subfloat[Difference between first 5, full mean, last 5, and last value]{\includegraphics[width=3in]{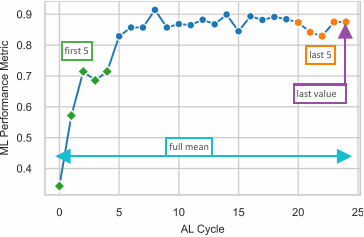}%
\label{fig:auc_schaubild1}}
\hfil
\subfloat[Difference between the ramp-up and the plateau phase]{\includegraphics[width=3in]{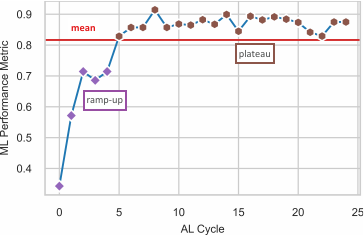}%
\label{fig:auc_schaubild2}}
\caption{Illustrating the different aggregation metrics}
\label{fig:auc_schaubild}
\end{figure*}
In terms of evaluation metrics, in addition to standard \gls{ML} metrics like accuracy, F1-score, precision, and recall, we incorporated aggregation metrics to escape reliance on learning curve plots, as illustrated in Fig.~\ref{fig:auc_schaubild}. First, we used the arithmetic mean over the whole curve (the \textit{full mean}). Then, we included variations to distinguish between the \textit{ramp-up} phase (early \gls{AL} cycles) and the \textit{plateau}-phase (later cycles). We used the average of the \textit{first} and \textit{last} 5 values, as well as only the \textit{last value}. Inspired by~\cite{reitmaierLetUsKnow2013, kottkeChallengesReliableRealistic2017}, we calculated a dataset-dependent dynamic threshold to distinguish between the ramp-up and plateau phases. Using the random \gls{AL} strategy as a baseline, approximated the plateau using the random strategy's mean performance across all cycles. This threshold is drawn as a single red line in Fig.~\ref{fig:auc_schaubild2}. If the assumption holds that there is a plateau to reach during the maximum \gls{AL} budget and that the plateau is much higher than the average in the early phase, the mean should be reasonably close to it. We then calculate a sliding window of 5 \gls{AL} cycles from the back of the timeseries, and the first time the mean of this sliding window is below the threshold, we differentiate between the ramp-up and the plateau phase.

Lastly, we also stored as the \textit{selected data samples}-metric per \gls{AL} cycle the indices of the currently selected batch $\unlabeledQuery$ for labeling to compare strategies on a more qualitative level compared to the high-level view from metrics.

\section{HPC experiments - implementation details}
\label{sec:implementation}
Running a series of \gls{AL} experiments $\ALExperimentSet$ over a large hyperparameter grid $\ALDatasetSet \times \ALTrainTestSplitSet \times \ALStartSetSet \times \ALStrategySet \times \batchSizeSet \times \learnerSet$ is not trivial, particularly, when one \gls{ML} model has to be trained in each \gls{AL} cycle. In this Section, we will share lessons learned from the practical implementation, which we believe will benefit other researchers who wish to reproduce or build upon our work.

To enhance reproducibility, we are making our source code fully available on GitHub~\footnote{\url{https://github.com/jgonsior/olympic-games-of-active-learning}}, and publish our complete raw experiment results on the data archive repository OPARA so other researchers can reuse and build upon our work~\footnote{\url{https://doi.org/10.25532/OPARA-862}}. In the following, we will explain our design decision for our code base.

The good news about \gls{AL} experiments is that they are highly parallelizable. We used a high-performance computing (HPC) cluster, as running the experiment on commodity hardware is not feasible. Our experiments consist of two stages: first, running the simulations, and second, analyzing the results. During the simulations, the resulting data grows to multiple terabytes in size. In the parallel setting, we found that appending results to CSV files worked best, as the order of concurrent appending write requests is irrelevant. A bottleneck is filesystem overloading and write collisions on single CSV files. We used a partitioning strategy and split our results by dataset and \gls{AL} strategy. This worked well with up to 10,000 parallel jobs, although we would recommend further partitioning (e.g., by a third and fourth hyperparameter besides dataset and strategy) for even larger scales. Most file-based compression algorithms are incompatible with append operations, so we did not compress files during the simulation phase.

Our hyperparameter grid of 4.6 million combinations required approximately 3.6 million CPU hours. For some hyperparameter combinations (e.g., QUIRE on the largest datasets), the runtime was multiple days, which was infeasible given the human-in-the-loop nature of \gls{AL}. Therefore, we enforced a runtime limit of 5 minutes per \gls{AL} cycle~\footnote{The anecdotal reason is that annotators are willing to wait as long as a typical coffee break is, which roughly equals 5 minutes}. Any hyperparameter combination exceeding this limit was excluded. Other works have also noted the need for runtime limits, with values ranging from 3 minutes to 7 days~\cite{margrafALPBenchBenchmarkActive2024, gonsiorImitALLearnedActive2022}.

It is not recommended to keep the results spread across multiple CSV files (due to our partitioning) during the analysis phase. We suggest merging all metrics into a single file in a column-store format, where each row contains all hyperparameters and the corresponding metric values. We found Apache Parquet to be the most effective format for storing this type of data. The resulting parquet files ranged from 60MiB to 6GiB. Using these files, all the analysis in the upcoming Evaluation Sec.~\ref{sec:evaluation} took only a few hours per plot. Caching intermediate results proved helpful when grouping results based on certain hyperparameters, as these can be reused in multiple evaluations.
\section{Evaluation}
\label{sec:evaluation}
Our empirical evaluation focuses on two major research questions:
\begin{description}
    \item[RQ1] What influence do the hyperparameters have on the results of an \gls{AL} experiment?
    \item[RQ2] Which hyperparameters should be included in a reliable and comparable \gls{AL} evaluation?
\end{description}
We give a brief overview of the data used in our experiments in Sec.~\ref{sec:dense_sparse}. For RQ1, we present two methods for quantifying the influence of a single hyperparameter in Sec.~\ref{sec:quantifying_influence}: first, \textit{metric-based correlation heatmaps} (Sec.~\ref{sec:metric-based-heatmaps}), and second, \textit{queried samples-based heatmaps} (Sec.~\ref{sec:queried-sample-based-heatmaps}). The results for individual hyperparameters are discussed in Sec.~\ref{sec:metrics} through Sec.~\ref{sec:runtime}.
For RQ2, we use \textit{leaderboard ranking invariance-based heatmaps} in Sec.~\ref{sec:leaderboard-ranking-invariance-heatmaps}. Finally, we summarize our results and provide recommendations in Sec.~\ref{sec:final_leaderboard}, with a conclusion on how to establish an empirical \gls{AL} setup with a hyperparameter set, which makes the results reproducible and comparable. %

The sections on RQ1 offer detailed findings on the influence of specific hyperparameter values. In contrast, the sections on RQ2 will provide high-level recommendations for setting up complete hyperparameter grids for \gls{AL} experiments.
\subsection{Completeness of the experimental results}
\label{sec:dense_sparse}
Our Hyperparameter grid $\ALExperimentSet$ consisted of exactly 4,636,800 hyperparameter combinations. Due to the 5-minute runtime limit per \gls{AL} cycle, we only exclude a neglectable amount of 75,924 combinations leaving us with 4,560,876 completed parameter combinations. These missing results are mainly from the largest datasets and both implementations of the QUIRE \gls{AL} strategy. Additionally, due to various errors in the implementations of \gls{AL} strategies (See f.e. \cite{luReBenchmarkingPoolBasedActive2023} for a detailed analysis of some implementation errors existent in the available open-source \gls{AL} frameworks), there are some missing results for almost all strategies and datasets. These errors occurred randomly across the entire parameter grid, and we will present an interpolation strategy in the following Sec.~\ref{sec:quantifying_influence} to handle these gaps in our nearly dense result grid.
\subsection{Quantifying the Influence of a Single Hyperparameter}
\label{sec:quantifying_influence}
Quantifying the influence of a single hyperparameter is not a straightforward task. We present three approaches: two methods based on the correlation between different hyperparameter combinations and one based on the correlation between \textit{leaderboard} rankings. The differences between the three methods lie in the aspect of correlation that is measured. High correlation indicates redundant hyperparameter values in an experimental hyperparameter grid.
\begin{itemize}
    \item \textbf{Metric-based heatmaps} (coloured in blue) measure the correlation between \gls{ML} metrics for two hyperparameter combinations. If two combinations produce similar \gls{ML} metrics, they have a high correlation.
    \item \textbf{Queried samples-based heatmaps} (colored in green) focus on the correlation between the sets of queried samples for labeling by the \gls{AL} strategies. High correlation means the same samples were selected for labeling.
    \item \textbf{Leaderboard ranking invariance-based heatmaps} (colored in orange) focus on comparing the rankings of \gls{AL} strategies between different hyperparameter grids. They correlate highly if two hyperparameter combinations produce similar rankings across strategies and datasets.
\end{itemize}
These three correlations do not necessarily agree, as discussed in the upcoming results Sections.
\subsubsection{Metric-based heatmaps}
\label{sec:metric-based-heatmaps}
The first method for quantifying hyperparameter influence relies on the extensive hyperparameter grid. We compare the results of a single metric between two hyperparameter combinations, with all parameters except for a single varying one, fixed. If there is a significant difference in the metric (e.g., average accuracy across all \gls{AL} cycles), this difference is attributed to the specific varying hyperparameter being evaluated. 

We aggregate the differences across all possible hyperparameters to get a quantitative idea of how much each hyperparameter influences \gls{AL} results. This approach works for all hyperparameters, except the initial start sets $\ALStartSetSet$ and the train-test-splits $\ALTrainTestSplitSet$, as these are not directly comparable between different datasets. A batch size of 5 is the same batch size of 5 for two different datasets, whereas a train-test-split based on a random seed is a different split on another dataset. These two parameters will be quantified later in Sec.~\ref{sec:leaderboard}.

We want to quantify the influence of a single hyperparameter; for illustration, we are using the batch size hyperparameter $\batchSizeSet$. First, a single vector $\ALFingerprintVector_{\batchSize_i}^T = [\ALExperimentResult | \ALExperimentResult \in \ALExperimentResultsSet_{\ALDatasetSet \times \ALTrainTestSplitSet \times \ALStrategySet \times \ALStartSetSet \times \learnerSet \times \batchSize} ],  \batchSize_i \in \batchSizeSet$ gets constructed for each $\batchSize_i \in \batchSizeSet$. $\ALFingerprintVector_{\batchSize_i}$ contains all the results from the experiments, having batch size $\batchSize_i$ as a hyperparameter. 
The vectors, e.g., $\ALFingerprintVector_{\batchSize_1}$ for $\batchSize_1$ and $\ALFingerprintVector_{\batchSize_2}$ for $\batchSize_2$, share a specific property: each dimension contains the results for the same set of hyperparameters, except the batch size. Thus, the experimental conditions for each dimension are equal, except for the batch size change, being either  $\batchSize_1$ or $\batchSize_2$.  We denote the result of the $j$-th hyperparameter combination having the hyperparameter $\batchSize_i$ and the metric $\ALMetric$ using $\ALMetric_{\batchSize{_i}j}$. We are using an aggregation metric, as the individual metric results per \gls{AL} cycle are too chaotic to compare them directly~\footnote{To get an idea of the chaotic learning curve plots, take a look at the variance in Fig.~\ref{fig:single_learning_curve_real}. The plot contains the results for 28 different hyperparameter values (28 strategies) for a single metric. Comparing each \gls{AL} cycle individually is not practical, as we do not expect each hyperparameter combination to get to the same metric value after $x$ \gls{AL} cycles, but the trend is similar. This can be achieved using the aggregation metrics.}:
\begin{gather*}
\ALFingerprintVector_{\batchSize_1}(\ALMetric) = \begin{bmatrix}
        \ALMetric_{\batchSize_11}\\
        \ALMetric_{\batchSize_12}\\
        \vdots
    \end{bmatrix}, 
    \ALFingerprintVector_{\batchSize_2}(\ALMetric)=\begin{bmatrix}
        \ALMetric_{\batchSize_21}\\
        \ALMetric_{\batchSize_22}\\
        \vdots
    \end{bmatrix}
\end{gather*}

From the result vector for a specific metric $\ALExperimentResult(\ALMetric)$, we can calculate how similar or dissimilar the outcomes of two \gls{AL} experiment are if only a single hyperparameter is changed, using the Pearson correlation coefficient $\pearson$~\cite{Pearson1895}. In the following Sections, we will display \textit{metric-based heatmaps}, containing the Pearson correlation coefficients between all combinations for a single variable hyperparameter (in this exemplary case: $\batchSizeSet = [\batchSize_1, …, \batchSize_K]$:
\begin{gather*}
\begin{split}
\begin{bmatrix} 
        \pearson(\ALFingerprintVector_{\batchSize_1}(\ALMetric), \ALFingerprintVector_{\batchSize_1}(\ALMetric)) & \cdots & \pearson(\ALFingerprintVector_{\batchSize_1}(\ALMetric), \ALFingerprintVector_{\batchSize_K}(\ALMetric))  \\
        \vdots &  \ddots & \vdots \\
       \pearson(\ALFingerprintVector_{\batchSize_K}(\ALMetric), \ALFingerprintVector_{\batchSize_1}(\ALMetric))&   \cdots     & \pearson(\ALFingerprintVector_{\batchSize_K}(\ALMetric), \ALFingerprintVector_{\batchSize_K}(\ALMetric))
    \end{bmatrix}
\end{split}
\end{gather*}
For better recognition, we are using a different color scheme for our heatmaps. The color for the metric-based heatmaps is blue.
\subsubsection{Queried samples-based heatmaps}
\label{sec:queried-sample-based-heatmaps}
Another approach to quantify the influence of a single hyperparameter focuses on the queried samples, instead of using an \gls{ML} metric as before. We use the list of samples, queried in each \gls{AL} cycle by the \gls{AL} strategy for labeling by the oracle $\ALExperimentResult(\unlabeledQuery)=[\unlabeledQuery^0, …, \unlabeledQuery^\AmountOfTrainingALCycles]$. Here, the vectors do not consist of numeric values (as before) but instead sets of sets of data samples. The resulting vectors, which represent the effect of a single hyperparameter in the parameter grid are therefore (in this example for the batch size hyperparameter values $\batchSize_1$ and $\batchSize_2$):
\begin{gather*}
    \ALFingerprintVector_{\batchSize_1}(\unlabeledQuery) = \begin{bmatrix}
    \widehat{\unlabeledQuery_{\batchSize_11}} = \bigcup_{i=0}^\AmountOfTrainingALCycles \unlabeledQuery_{\batchSize_{1}1}^i\\
    \widehat{\unlabeledQuery_{\batchSize_12}} = \bigcup_{i=0}^\AmountOfTrainingALCycles \unlabeledQuery_{\batchSize_{2}2}^i\\
        \vdots
    \end{bmatrix}\\
    \ALFingerprintVector_{\batchSize_2}(\unlabeledQuery)=\begin{bmatrix}
    \widehat{\unlabeledQuery_{\batchSize_21}} = \bigcup_{i=0}^\AmountOfTrainingALCycles \unlabeledQuery_{\batchSize_{2}1}^i\\
    \widehat{\unlabeledQuery_{\batchSize_22}} = \bigcup_{i=0}^\AmountOfTrainingALCycles \unlabeledQuery_{\batchSize_{2}2}^i\\
        \vdots
    \end{bmatrix}
\end{gather*}

Analog to the vectors from the metric results, each dimension contains the results for the same combinations of hyperparameters, except for the different hyperparameters of, e.g., the batch size $\batchSize$. We are denoting the query $\unlabeledQuery$ of the first hyperparameter combination having $\batchSize_1$ as batch size with $\unlabeledQuery_{\batchSize_{1}1}$. It is not relevant which hyperparameter is exactly the first one, it is only important that the order is consistent across the vectors $\ALFingerprintVector$, to ensure comparability between the vector dimensions.

Since we cannot directly compute the Pearson correlation for vectors of  sets of sets, we are first taking the concatenated set of labeled samples  $\hat{\unlabeledQuery} = \bigcup_{i=0}^\AmountOfTrainingALCycles \unlabeledQuery^i$ as elements in the vector $\ALFingerprintVector$, and second use the Jaccard index~$J$, to calculate dimension for dimension for two given vectors $\ALFingerprintVector_{\batchSize_1}(\unlabeledQuery)$ and $\ALFingerprintVector_{\batchSize_2}(\unlabeledQuery)$, how similar the queried sets for each iteration are. The Jaccard index~\cite{murphyFinleyAffairSignal1996} proved to generate more sound results than rank correlation measures such as Kendall Tau~\cite{kendallNewMeasureRank1938} or Spearman's rank correlation coefficient~\cite{spearmanProofMeasurementAssociation1904}. This can be attributed to the high fluctuations in \gls{AL} cycles, often resulting in two very different rankings despite a high Jaccard similarity. For a given pair of hyperparameter values, for example $\batchSize_1$ and $\batchSize_2$, the jaccard vector $\ALFingerprintVector_{J_{\batchSize_1\batchSize_2}}$ is computed:
\begin{gather*}
    \ALFingerprintVector_{J_{\batchSize_1\batchSize_2}} = \begin{bmatrix}
        J(\widehat{\unlabeledQuery_{\batchSize_11}}, \widehat{\unlabeledQuery_{\batchSize_21}})\\
        J(\widehat{\unlabeledQuery_{\batchSize_12}}, \widehat{\unlabeledQuery_{\batchSize_22}})\\
        \vdots
    \end{bmatrix}
\end{gather*}

As the last step, we take the sum of the resulting jaccard vector $\ALFingerprintVector_{J_{\batchSize_1\batchSize_2}}$, and divide it by the length of the respective vector to be able to generate the exemplary heatmap for the batch size hyperparameter $\batchSizeSet = [\batchSize_1, …, \batchSize_K]$. For consistency with the metric-based heatmaps, we subtract the Jaccard index from 1 so that 1 indicates full similarity. The heatmaps for \textit{queried samples-based heatmap} are colored in green:%
\begin{gather*}
    \begin{split}
\begin{bmatrix} 
        1-\frac{\sum_{j \in \ALFingerprintVector_{J_{\batchSize_1\batchSize_1}}}j}{|\ALFingerprintVector_{J_{\batchSize_1\batchSize_1}}|}=1 & \cdots & 1-\frac{\sum_{j \in \ALFingerprintVector_{J_{\batchSize_1\batchSize_K}}j}}{|\ALFingerprintVector_{J_{\batchSize_1\batchSize_K}}|}  \\
        \vdots &  \ddots & \vdots \\
       1-\frac{\sum_{j \in \ALFingerprintVector_{J_{\batchSize_K\batchSize_1}}}j}{|\ALFingerprintVector_{J_{\batchSize_K\batchSize_1}}|}&   \cdots     & 1-\frac{\sum_{j \in \ALFingerprintVector_{J_{\batchSize_K\batchSize_K}}}j}{|\ALFingerprintVector_{J_{\batchSize_K\batchSize_K}}|}=1
\end{bmatrix}
\end{split}
\end{gather*}

\subsubsection{Leaderboard ranking invariance-based heatmaps}
\label{sec:leaderboard-ranking-invariance-heatmaps}
This approach is conceptually different from the previous two, as it focuses on comparing the rankings of \gls{AL} strategies under different hyperparameter grids. For this, we fix all hyperparameters except one and compare the rankings of \gls{AL} strategies across multiple datasets in a so-called \textit{leaderboard}.  Table~\ref{tab:example_leaderboard} shows an exemplary leaderboard. 

The content of each leaderboard cell is denoted as $\leaderboardCell_{\ALStrategy_i\ALDataset_j}$, which represents the metric results for a specific \gls{AL} strategy $\ALStrategy_i$ on a dataset $\ALDataset_j$, averaged over all fixed hyperparameter combinations (with the hyperparameter under analysis being the variable one): 
\begin{gather*}
    \leaderboardCell_{\ALStrategy_i\ALDataset_j} = \frac{\sum_{\ALMetric_{k} \in \ALFingerprintVector_{\ALDataset_i\ALStrategy_j}(\ALMetric)}\left(\ALMetric_{k}\right)}{|\ALFingerprintVector_{\ALDataset_i\ALStrategy_j}|}
\end{gather*}
\begin{table}[]
\centering
\begin{tabular}{lccr}
\toprule
          & Strategy $\ALStrategy_1$ & Strategy $\ALStrategy_2$ & … \\ \midrule
Dataset $\ALDataset_1$ &$\leaderboardCell_{\ALStrategy_1\ALDataset_1}$            &     $\leaderboardCell_{\ALStrategy_2\ALDataset_1}$        &  … \\
Dataset $\ALDataset_2$ &$\leaderboardCell_{\ALStrategy_1\ALDataset_2}$          &       $\leaderboardCell_{\ALStrategy_2\ALDataset_2}$      &  … \\
\vdots         &       \vdots     &    \vdots        &  \vdots \\ \midrule
Final Ranking $\finalRanking$     &     $\frac{\sum_{\ALDataset_i \in \ALDatasetSet}\leaderboardCell_{\ALStrategy_1\ALDataset_i}}{|\ALDatasetSet|}$      &  $\frac{\sum_{\ALDataset_i \in \ALDatasetSet}\leaderboardCell_{\ALStrategy_2\ALDataset_i}}{|\ALDatasetSet|}$  & … \\ 
\bottomrule
\end{tabular}
\caption{Leaderboard results}
\label{tab:example_leaderboard}
\end{table}
The leaderboard can then be reduced to a final ranking of all strategies. This final ranking can be a vector of the average results per strategy. If changing a hyperparameter significantly alters the rankings, this indicates the hyperparameter has a strong influence. To calculate correlations between the rankings, we use a  two-sided Kendall's tau-b rank correlation test~\cite{kendallNewMeasureRank1938}, which is robust to outliers and differences in ranking distributions.  \textit{Leaderboard ranking invariance heatmap} are colored in orange.

Using leaderboards to compare \gls{AL} strategies is not straightforward and comes with a few practical challenges. The most significant issue is the comparison of metrics results across datasets. For example, dataset $\ALDataset_1$ may have accuracy scores between 50\% and 70\%, while dataset $\ALDataset_2$ might have accuracy scores between 42\% and 45\%. In this case, \gls{AL} strategies optimized for dataset $\ALDataset_1$ would show more significant absolute improvements than those optimized for $\ALDataset_2$, even if they perform equally well in relative terms.

To overcome this, some researchers normalize the metric values across datasets or use \textit{ranks} instead of absolute values~\cite{kottkeChallengesReliableRealistic2017, gonsiorImitALLearnedActive2022}. We chose ranks as normalization would add a layer of complexity due to the selection of the normalization method (e.g., outlier resistance or not).

Another issue arose during our evaluation because of the sparseness of the result grid missing some hyperparameter combinations, as thus the leaderboard cell entries $\leaderboardCell$ are not all aggregated on the same amount of hyperparameter combinations. Therefore, to make the comparisons between the final ranking vectors even, we are interpolating the results using a fixed value of $0$, meaning the worst possible outcome for the \gls{ML} metric. This punishes strategies that either take longer than the runtime limit or result in runtime errors.
\subsection{Machine learning metrics performance}
\label{sec:metrics}
\begin{figure*}[!htp]
\centering
\subfloat[Metric-based heatmap, basic ML metrics]{\includegraphics[width=3in]{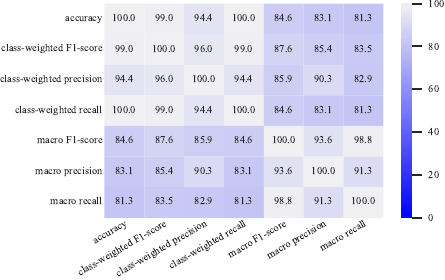}%
\label{fig:heatmap_standard_ml_metric}}
\hfil
\subfloat[Metric-based heatmap, aggregation ML metrics]{\includegraphics[width=3in]{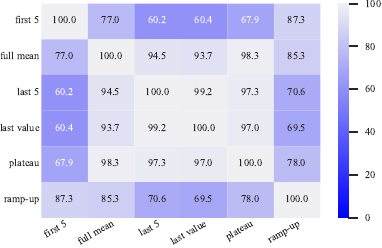}%
\label{fig:heatmap_aggregtaion}}
\hfil
\subfloat[Leaderboard ranking-invariance heatmap, basic ML metrics]{\includegraphics[width=3in]{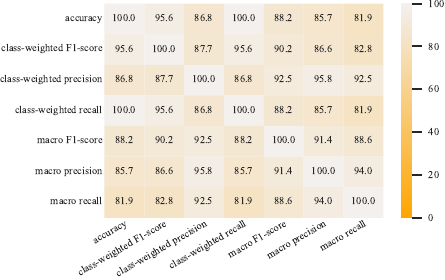}%
\label{fig:leaderboard_basic_metric}}
\hfil
\subfloat[Leaderboard ranking-invariance heatmap, aggregation ML metrics]{\includegraphics[width=3in]{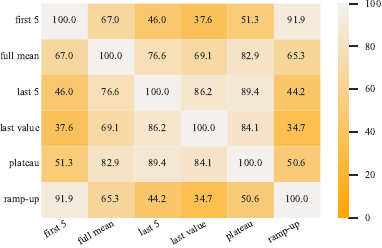}%
\label{fig:leaderboard_auc_metrics}}
\caption{Metric-based and leaderboard ranking invariance heatmaps for the basic as well as the aggregation metrics}
\label{fig:metric_heatmaps}
\end{figure*}
The first hyperparameter we investigate is the influence of the \gls{ML} metric. This parameter is evaluated first, as the results determine which metric $\ALMetric$ we will use in the upcoming evaluations for the vectors $\ALFingerprintVector$. The hyperparameter metric has two aspects: first, the foundational \gls{ML} metric (like accuracy), which measures the performance of the learner model for each \gls{AL} cycle, resulting in a timeseries; and second, the aggregation metric, which compacts the learning curve or timeseries into a single value.

For both aspects, we provide two plots. First, a heatmap using the quantification method based on the \gls{ML} metric results and the Pearson correlation coefficient. Second, a heatmap based on the correlation between the final leaderboard rankings. Fig.~\ref{fig:heatmap_standard_ml_metric} and Fig.~\ref{fig:heatmap_aggregtaion} show the Pearson correlation heatmaps for the foundational and aggregation metrics, while Fig.~\ref{fig:leaderboard_basic_metric} and Fig.~\ref{fig:leaderboard_auc_metrics} display the leaderboard heatmaps. The heatmaps present the Pearson correlation coefficient between all results in our hyperparameter grid, comparing one metric with another. High values close to 100 indicate that both metrics behave similarly, while low values close to 0 indicate no correlation. As the queried samples are the same for each metric, we do not present queried samples-based heatmaps for this part.

For the evaluations in the upcoming sections, a single \gls{ML} metric is preferred. Since it is challenging to determine the best metric, we focus on the one with the highest correlation to all other metrics. This ensures that results based on this metric will be representative of the other metrics as well. Unsurprisingly, the F1-score, which is the harmonic mean of both precision and recall, shows the highest correlation. Thus, for the upcoming Sections, we use the the F1-score  as metric $\ALMetric$, since it correlates highly with the accuracy metric and is better suited for datasets with many classes and class imbalances. The results are fairly similar between the metric-based and the leaderboard ranking invariance heatmap.

Regarding the aggregation metrics displayed in Fig.~\ref{fig:heatmap_aggregtaion} and Fig.~\ref{fig:leaderboard_auc_metrics} (see Sec.~\ref{sec:our_param_grid} and Fig.~\ref{fig:auc_schaubild} for details about the included aggregation metrics), it is not surprising that there is a strong correlation between the first 5 values and the ramp-up phase, as well as between the last 5, the last value, and the plateau phase. This pattern holds in both the leaderboard ranking-invariance heatmap and the metric-based heatmap. 

There are arguments for a preference for both phases in an evaluation: there is more fluctuation in the ramp-up phase, and the results only stabilize in the plateau phase, which would be an argument for using plateau-phase metrics. On the other hand, most gains in \gls{AL} are in the ramp-up phase, as the plateaus are often very similar, which is not surprising, as a certain amount of labeled data will always turn out in very similar results. As the full mean metrics cover both phases and correlate highly with all aggregation metrics, we will safely use this metric in the upcoming evaluations. The differences in the metrics become more obvious in the leaderboard ranking invariance heatmap, even though the results in both heatmaps are quite similar. 

As the correlations vary drastically for the aggregation metric, we advise to be cautious when deciding on an aggregation metric apart from the full mean. Especially given the unclear distinction between a ramp-up and a plateau phase in the context of \gls{AL}, we do not recommend using either phase as the sole evaluation focus.
\subsection{Batch size}
\begin{figure*}[!htp]
\centering
\subfloat[Metric-based]{\includegraphics[width=2in]{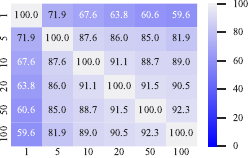}%
\label{fig:batch_metric}}
\hfil
\subfloat[Queried samples-based]{\includegraphics[width=2in]{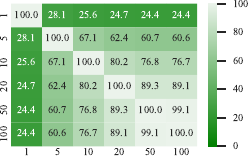}%
\label{fig:batch_indices}}
\hfil
\subfloat[Leaderboard ranking invariance-based]{\includegraphics[width=2.5in]{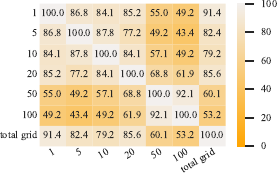}%
\label{fig:batch_leaderboard}}
\caption{Heatmaps for the batch size hyperparameter $\batchSizeSet$}
\label{fig:batch_heatmaps}
\end{figure*}
The results for the batch size hyperparameter, shown in Fig.~\ref{fig:batch_heatmaps}, are as expected. Regardless of the heatmap used (metric-based, queried samples-based, or leaderboard ranking invariance-based), there is a noticeable trend: batch sizes correlate highly when their values are close. Smaller batch sizes tend to behave similarly, and the same goes for large batch sizes. However, large batch sizes differ significantly from small ones. This is most evident in the queried samples-based heatmaps. Since most \gls{AL} strategies select batches based on a top-$k$ ranking, the top-$50$ samples are expected to be a substantial subset of the top-$100$ samples, explaining the high correlation of $99.1\%$. As different batch sizes correlate higher with each other for the metric- or leaderboard ranking invariance-based heatmaps it does not matter so much which exact samples are queried to get the same results. Especially for large datasets, many samples are likely equivalent, leading to similar leaderboard rankings despite differences in the queried samples.

The correlation to the entire parameter grid was included as an additional row/column in the leaderboard ranking invariance-based heatmap. This shows that very large batch sizes result in vastly different rankings, whereas lower values mostly agree.

Overall, we conclude that in an \gls{AL} experimental evaluation, it is advisable to include at least two batch sizes: one very small and one larger value. Using two large batch sizes is not recommended, as the results would be too similar. If only a single batch size can be included, a batch size of 20 showed the highest correlation to all other values on average.%
\subsection{Datasets}
Unfortunately, the heatmap visualizations for the dataset hyperparameter, which spans 92 dimensions, are too large to be displayed in this paper. However, they can be found in the GitHub repository associated with this study~\footnote{\url{https://github.com/jgonsior/olympic-games-of-active-learning/tree/main/plots}}. These heatmaps confirm the findings of other works, showing that the dataset parameter significantly impacts the performance of \gls{AL} strategies. Unlike the other hyperparameters, datasets exhibit substantial variation, with correlation values ranging from as low as 30\% to as high as 100\%. Given this variability, we do not recommend exclusively using a specific set of datasets. Instead, we advise using an extensive, diverse collection of datasets, such as the well-known OpenML-CC18 benchmark suite~\cite{bischlOpenMLBenchmarkingSuites2021}, which includes datasets with various characteristics and levels of complexity.
\subsection{Learner models}
\begin{figure*}[!htp]
\centering
\subfloat[Metric-based]{\includegraphics[width=1.5in]{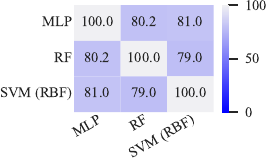}%
\label{fig:learner_metric}}
\hfil
\subfloat[Queried samples-based]{\includegraphics[width=1.5in]{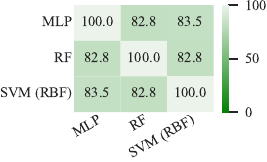}%
\label{fig:learner_jaccard}}
\hfil
\subfloat[Leaderboard ranking invariance-based]{\includegraphics[width=1.5in]{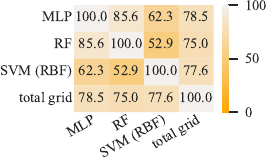}%
\label{fig:learner_leaderboard}}
\caption{Heatmaps for the learner model hyperparameter $\learnerSet$}
\label{fig:learner_model_heatmaps}
\end{figure*}
The hyperparameter of the \gls{ML} learner model $\learner$ shows that our three compared model implementations behave very similarly. The metric- and queried samples-based heatmaps show very high correlations between the included hyperparameter values in our experimental grid. The results for the leaderboard ranking invariance-based heatmap, displayed in Fig.~\ref{fig:learner_model_heatmaps} are slightly different: even though the RF model behaves still very close to the MLP model, the SVM (RBF) model results in very different final rankings. The correlation between the RF model and the SVM (RBF) model is especially low. Our explanation lies in the different correlations measured in the three heatmaps: a high correlation in the queried samples-based heatmap, and a similar high correlation in the metric-based heatmap shows that similar samples are selected for labeling and that the respective reached \gls{ML} metrics correlate with each other. As the leaderboard ranking invariance-based heatmap shows less correlation, the similar selected samples and similar reached \gls{ML} metric results do not result in similar final leaderboard rankings. Slight changes in the underlying metric can easily cause a ranking change. Potentially, these are the remaining $~20\%$ not present in the correlation in the other two heatmaps.
Regarding the question of why especially the RF and SVM (RBF) models behave so differently compared to the MLP model, we hypothesize that both models have a disjunct set of datasets where they perform very well, potentially datasets with many classes for the RF model, and datasets which are sparse for the SVM model. As the MLP model can adapt to both settings very well, it can correlate better with both models. The last row in the leaderboard ranking invariance-based heatmap shows the correlation between the experiment grids limited to the respective single learner model compared to the total grid. As the MLP model has the highest correlation to the complete grid of all results, this also backs up our hypothesis that the MLP model can perform well for both datasets and is better suited to the RF and the SVM (RBF) models.

In conclusion, we advise using the MLP model, which seems to produce the same results as the RF and SVM (RBF) models.
\subsection{Active learning query strategies}
\begin{figure*}[!htp]
\centering
\subfloat[Metric-based heatmap]{\includegraphics[width=.92\textwidth]{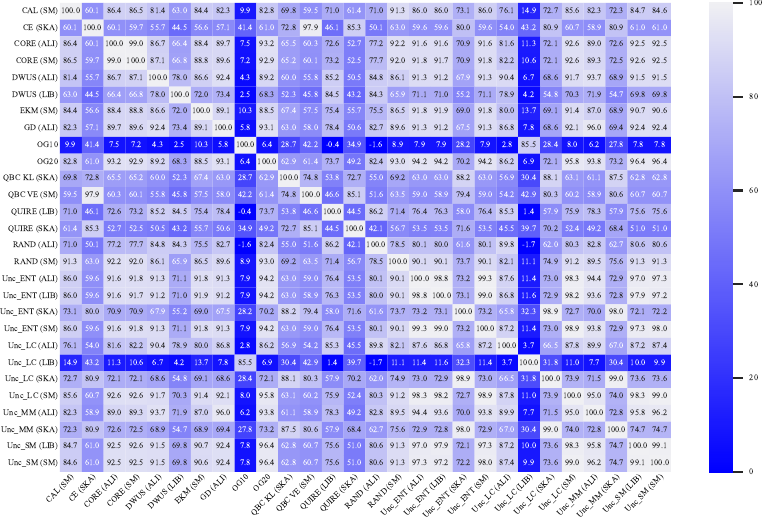}%
\label{fig:strategy_metric}}
\hfil
\subfloat[Queried samples-based heatmaps]{\includegraphics[width=.92\textwidth]{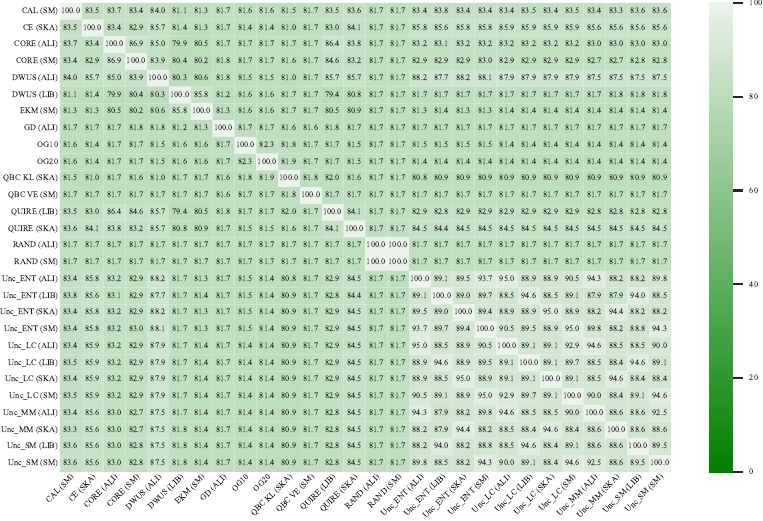}%
\label{fig:strategy_jaccard}}
\caption{Heatmaps for the \gls{AL} query strategy hyperparameter \gls{AL} strategy $\ALStrategySet$}
\label{fig:strategy_param}
\end{figure*}
Our experimental grid includes 28 \gls{AL} strategy implementations from various frameworks and authors. The leaderboard ranking invariance-based heatmap depends on the presence of multiple \gls{AL} strategies and, therefore, cannot measure the correlation between two strategies like the other hyperparameters. Thus, Fig.~\ref{fig:strategy_param} presents the metric-based heatmap (a) and the queried samples-based heatmap (b).

We begin by interpreting the metric-based correlations, which show stronger correlations and provide more valuable insights. We anticipated that the same strategy would perform similarly when implemented across different frameworks. Notable strategies with multiple implementations include CORE, DWUS, QUIRE, RAND, and all four uncertainty variants (ENT, LC, MM, and SM). Additionally, QBC and OG are each implemented in two versions. As expected, the two CORE variants behave almost identically, with a $99.0\%$ correlation. The two DWUS variants show some similarities but are not as pronounced compared to their correlations with other strategies. Interestingly, the two QUIRE implementations exhibit a relatively low correlation of only $44.5\%$, indicating significant differences.

The most notable finding is the high correlation among all direct uncertainty-based variants, with a few exceptions. Unc\_ENT implemented in SKA, Unc\_LC implemented in LIB, and Unc\_MM implemented in SKA are outliers. The three variants of Unc\_LC, Unc\_MM, and Unc\_Ent in the framework SKA are more closely related to each other than to implementations of the same strategy in the different frameworks. Surprisingly, we conclude that the specific implementation of \gls{AL} strategies can impact performance more than the underlying strategy itself.
\begin{figure}[!htp]
    \centering
    \includegraphics[width=2in]{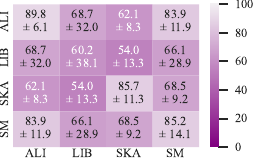}
    \caption{Average correlation and standard deviation among the \gls{AL} strategies between the frameworks}
    \label{fig:al_strat_framework_corr}
\end{figure}
To further investigate the influence of the framework on \gls{AL} strategy implementation, we calculated the average correlation among strategies within each framework, including standard deviations. As shown in Fig.~\ref{fig:al_strat_framework_corr}, there are no high correlations observable between framworks, except for ALI and SM. However, given the high standard deviation, these correlations are likely due to a few similarly implemented strategies. Interestingly, we observed a trend of high correlation among strategies within the same framework despite the variety of strategies included in each framework. The self-correlation of $89.8\%$ for ALI is exceptionally high, while at the same time the standard deviation is the lowest overall. This leads us again to the conclusion that \gls{AL} strategies have very high degrees of freedom in their implementation and that some design choices inside of a framework can have a very high impact on the performance of the \gls{AL} strategy, even further than the concrete \gls{AL} strategy. It is out of the scope of this paper to investigate in detail how the different implementations differ from each other. Still, we believe this is an interesting research question that should be addressed in future research endeavors.

\gls{AL} strategies can be categorized broadly into two categories~\cite{zhanComparativeSurveyBenchmarking2021}, uncertainty-based~\cite{lewisSequentialAlgorithmTraining1994,schefferMiningWebActive2001, shannonMathematicalTheoryCommunication1948, schroderSurveyActiveLearning2020, zhangCartographyActiveLearning2021, kirschBatchBALDEfficientDiverse2019,houlsbyBayesianActiveLearning2011, siddhantDeepBayesianActive2018}, and diversity-based strategies~\cite{colemanSelectionProxyEfficient2020,senerActiveLearningConvolutional2018}. Modern strategies aim to combine both~\cite{hsuActiveLearningLearning2015, konyushkovaDiscoveringGeneralPurposeActive2019, konyushkovaLearningActiveLearning2017, huangActiveLearningQuerying2010, gonsiorImitALLearnedActive2022, wangQueryingDiscriminativeRepresentative2015}. The uncertainty-based strategies in this paper are CAL, CE, QBC, and all Unc variants (ENT, LC, MM, and SM), the diversity-based are CORE, EKM, and GD, and the combined variants are DWUS and QUIRE. As baselines serve the random strategy, as well as the optimal strategy OG, capable of \textit{peeking} into the future to make a greedy near-optimal decision based on the knowledge of what the true labels are (which is, of course, knowledge only present in \gls{AL} experiments, and never present in real-world settings). We could not find any proof in our experimental results showing a general similarity among all the \gls{AL} strategies of a similar category, neither uncertainty-based nor diversity-based, nor for the combined strategies.

We included the near-optimal baseline strategies OG10 and OG20 as an upper barrier to measuring how good current implementations are. OG10 has access to the future for 10 queries, whereas OG20 has access to 20 queries. Interestingly, this difference significantly affects how similar both strategies are to all others. OG10 is dissimilar to almost everything, whereas OG20 highly correlates to many strategies. We explain that there is indeed a shared property that marks some data samples as \textit{good} for \gls{AL}, and OG20 can find it due to the knowledge of more future foresight, compared to OG10, which with a comparably small foresight mostly behaves randomly. We will come back later to this point in Sec.~\ref{sec:final_leaderboard}, where we discuss if the similarity to OG20 hints at an overall good performing \gls{AL} strategy.

Many correlations in the metric-based heatmap become non-existent in the queried samples-based heatmap. This can be easily explained by the fact that multiple data samples often have the same training outcome as they are very similar/close to each other but still have different samples. Small implementation details such as different sorting, less than vs. less than or equal, etc., can make a huge difference. Given that most correlations here are at least $80\%$ or higher, it indicates that most strategies select the same samples overall. Still, their ordering is what makes the real difference regarding performance. Again, most Unc strategies are close to each other, without any noteworthy exceptions compared to the metric-based correlation; even the poorly metric-correlated Unc\_LC implementation in LIB is now closely similar to all other strategies.

The correlation of the random baseline strategies regarding the queried samples is $100.0\%$, which is very high compared to the metric-based correlation of $78.5\%$. This hints at the complexity of \gls{AL} overall. Even though both strategies selected the same number of samples for labeling in the end, the resulting \gls{ML} metrics are not always identical. We explain this discrepancy by different orderings in the selected samples at each iteration, which leads to a different order of the samples presented to the \gls{ML} learner models and, therefore, slightly different classification boundaries. This shows that the selection of the metric, either being a \gls{ML} metric, or the set of selected indices, leads to a very different interpretation of how similar or different two \gls{AL} experiments are.

\subsection{Train-test-split}
\begin{figure*}[!htp]
\centering
\subfloat[Metric-based]{\includegraphics[width=2in]{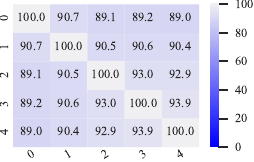}%
\label{fig:train_test_heatmaps_metric}}
\hfil
\subfloat[Leaderboard ranking
invariance-based]{\includegraphics[width=2in]{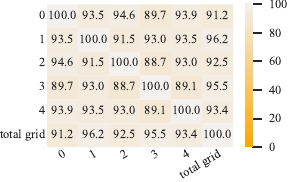}%
\label{fig:train_test_heatmaps_jaccard}}
\caption{Heatmaps for the hyperparameter train-test-split $\ALTrainTestSplitSet$}
\label{fig:train_test_heatmaps}
\end{figure*}
The hyperparameter train-test-split $\ALTrainTestSplitSet$ differs from the previously evaluated ones, as the concrete parameter values are incomparable across datasets. This hyperparameter describes a specific train-test split, e.g., the fourth train-test-split of dataset D1 differs from the fourth train-test-split of dataset D2. Therefore, the correlation heatmap plots do not tell us anything about the influence of a specific value for this parameter. However, they do tell us more about the general influence this parameter can have if changed. The queried samples-based heatmap is not possible, as due to different train-test-splits, the two possible sets of unlabeled samples between two splits are different. Fig.~\ref{fig:train_test_heatmaps} contains the metric-based heatmap (a; blue) as well as the leaderboard ranking
invariance-based heatmap (b; yellow). The correlations show that this hyperparameter has no considerable influence on the results of \gls{AL} experiments, as the results correlate both metric-wise and for the resulting final leaderboard rankings highly with each other. We will give an additional explanation of why this parameter seems unimportant in the later Sec.~\ref{sec:minimum_parameter} but may be more critical than these plots show.
\subsection{Start point sets}
\begin{figure*}[!htp]
\centering
\includegraphics[width=.8\linewidth]{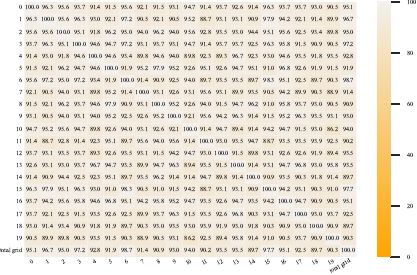}%
\caption{Leaderboard ranking
invariance-based heatmap for the hyperparameter start set $\ALTrainTestSplitSet$}
\label{fig:start_point_heatmaps}
\end{figure*}
The results for the hyperparameter of the start point set $\ALStartSetSet$ are displayed in Fig.~\ref{fig:start_point_heatmaps}. The results are the same as for the train-test-splits, overall this hyperparameter seems to have no big influence on the outcome of a \gls{AL} experiment. This is a surprising result, as one would expect that the choice of the initially labeled samples greatly influences the outcome. Why this is potentially still the case will be explained in Sec.~\ref{sec:minimum_parameter}.
\subsection{What is the minimum amount of needed parameters after all?}
\label{sec:minimum_parameter}
\begin{figure*}[!htp]
\centering
\subfloat[Total grid]{\includegraphics[width=2in]{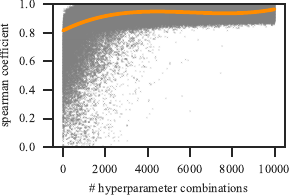}%
\label{fig:min_hyper2}}
\hfil
\subfloat[Without batch size $\batchSizeSet$]{\includegraphics[width=2in]{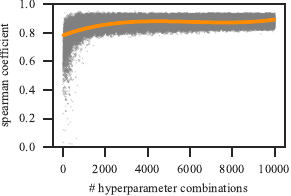}%
\label{fig:reduction_batch}}
\hfil
\subfloat[Without learner model $\learnerSet$]{\includegraphics[width=2in]{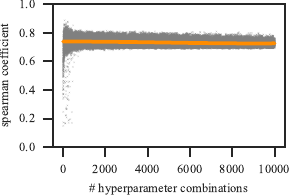}%
\label{fig:reduction_learner}}
\hfil
\subfloat[Without initially labeled start set $\ALStartSetSet$]{\includegraphics[width=2in]{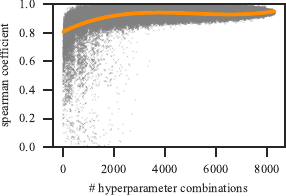}%
\label{fig:rudction_start_point}}
\hfil
\subfloat[Without train-test-split $\ALTrainTestSplitSet$]{\includegraphics[width=2in]{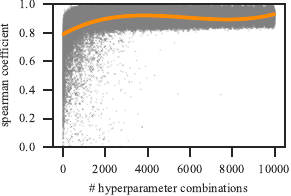}%
\label{fig:reduction_train_test_bucket}}
\caption{Hyperparameter combination amount simulations}
\label{fig:hyper_reduction}
\end{figure*}
Besides analyzing the influence of specific \gls{AL} hyperparameters, our research focuses on recommending which hyperparameter grid should be used in empirical evaluations, ensuring that the results are reproducible among \gls{AL} studies. We are looking for the minor grid of parameter values, which still reliably produces the same results as almost every other parameter combination. Using a large hyperparameter grid similar to ours is not practically feasible, as it exceeds the computational budgets of most research projects. Under the assumption that our hyperparameter grid is large enough to include almost any other possible \gls{AL} evaluation scenarios, a strong correlation to the complete grid's results should indicate a hyperparameter grid that fits our target criteria. As a correlation, we are using the leaderboard ranking-invariance.

Fig.~\ref{fig:min_hyper2} shows a scatterplot of the Spearman correlation coefficient between the total grid of hyperparameters used in this study and randomly sampled subsets of hyperparameter combinations of growing size. For each amount of hyperparameter combinations on the x-axis, we have drawn 100 random subsets. The orange line shows a polynomial regression of maximum order 3 to give a correlation trend. It can be seen that after below 2,000 random hyperparameter combinations, the correlations have a very high spread, indicating a high chance of picking a set of hyperparameters that will produce an experimental outcome that will be contrary to those of similarly sized hyperparameter sets. This is probably the leading cause of why the results of many \gls{AL} studies contradict each other about the superiority of single \gls{AL} strategies over another. Starting from around 2,000 combinations, though, the results seem to stabilize, and from around 4,000 combinations onwards, a strong correlation between 80\% and 100\% is almost guaranteed. So our first finding can be summarized as, given a wide variety in the hyperparameter grid (around 4.6 million combinations of 6 hyperparameters in our case), a fraction of at least 4,000 combinations is already enough to produce the same results as for the complete grid.

We hypothesize that the main quality the smaller subset of sufficient hyperparameter combinations must share is a wide variety of hyperparameters from which they are drawn. It does not matter which concrete hyperparameter values are included in the grid. It is much more important to have a large variety of hyperparameters. To further back up this assumption, we have removed each hyperparameter (except for the set of \gls{AL} strategies $\ALStrategySet$, and the set of datasets $\ALDataset$, which both have stayed the same for the leaderboard ranking-invariance correlations), and repeated the experiment of calculating the correlation between the smaller drawn subsets, and the total grid. The results are shown in Fig.~\ref{fig:reduction_batch} to Fig.~\ref{fig:reduction_train_test_bucket}. For example, by removing the batch size hyperparameter from the grid, we used instead of the complete grid of $\ALDatasetSet \times \ALTrainTestSplitSet \times \ALStartSetSet \times \ALStrategySet \times \batchSizeSet \times \learnerSet$ the smaller grid of $\ALDatasetSet \times \ALTrainTestSplitSet \times \ALStartSetSet \times \ALStrategySet \times \learnerSet$ and used a fixed batch size of 20. Thus, we reduce the \textit{entropy} in the possible parameter grid from which the smaller subset could be drawn.

The plots show that removing the hyperparameters of the batch size and the learner model substantially impacts the correlation between the randomly drawn subsets and the complete hyperparameter grid. The results fluctuate less for small hyperparameter samples, and the correlations generally have less variability. Also, a strong correlation of 95\% and above can never be reached after removing these two parameters completely from the grid. In contrast, removing the initially labeled start set and the train-test-split hyperparameter does not seem to impact the distribution of the correlations at all; they look almost identical. This is expected, as both hyperparameters depend on the dataset and the respective samples of a dataset. As the corresponding other hyperparameter and the dataset still exist in the hyperparameter grid, there is still enough entropy due to the dataset-related hyperparameters present. Therefore, these two are hyperparameters that can be safely removed from the grid, but do not harm anything if kept.
\subsection{Runtime}
\label{sec:runtime}
\begin{figure*}[!htp]
    \centering
    \includegraphics[width=\linewidth]{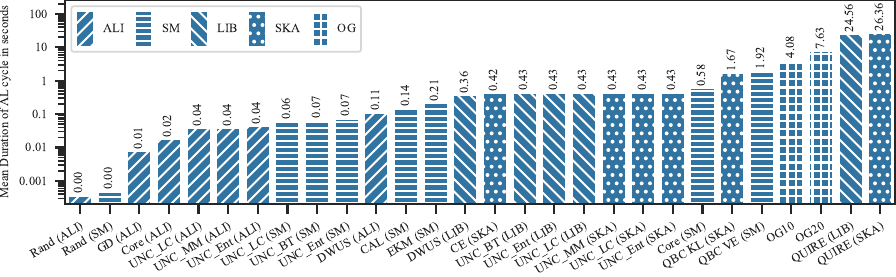}
    \caption{Average Runtime of all strategies over the complete hyperparameter grid, logarithmic scale}
    \label{fig:runtime}
\end{figure*}
The runtime performance of \gls{AL} strategies is rarely considered in \gls{AL} studies. \gls{AL} is a human-in-the-loop approach, where humans label the selected queries. As human labor is the main reason to use \gls{AL} in the first place, the fast runtime performance of \gls{AL} strategies is a crucial deciding factor in practice, with the goal of minimal idle human time. In Fig.~\ref{fig:runtime}, the average duration of a single \gls{AL} cycle, averaged over the complete hyperparameter grid, is shown for each \gls{AL} strategy, using a logarithmic scale. There is a noticeable difference between fast strategies, such as most uncertainty-based variants, and more complex strategies, such as Core, QBC, or QUIRE. Given that the majority of our datasets are small and our 5-minute runtime limit per \gls{AL} cycle, these differences will only become more evident for huge datasets, such as in the NLP or image domain. 
A clear distinction is noticeable between the strategies implemented in different frameworks. We hypothesize that the data model used inside the frameworks has the most prominent performance impact. We encourage other \gls{AL} researchers to consider runtime as a metric for evaluating \gls{AL} strategies in future work, in addition to \gls{ML} performance metrics, as it is crucial to keep annotators motivated.
\subsection{Final leaderboard}
\label{sec:final_leaderboard}
\begin{figure*}[!htp]
    \centering
    \includegraphics[width=\linewidth]{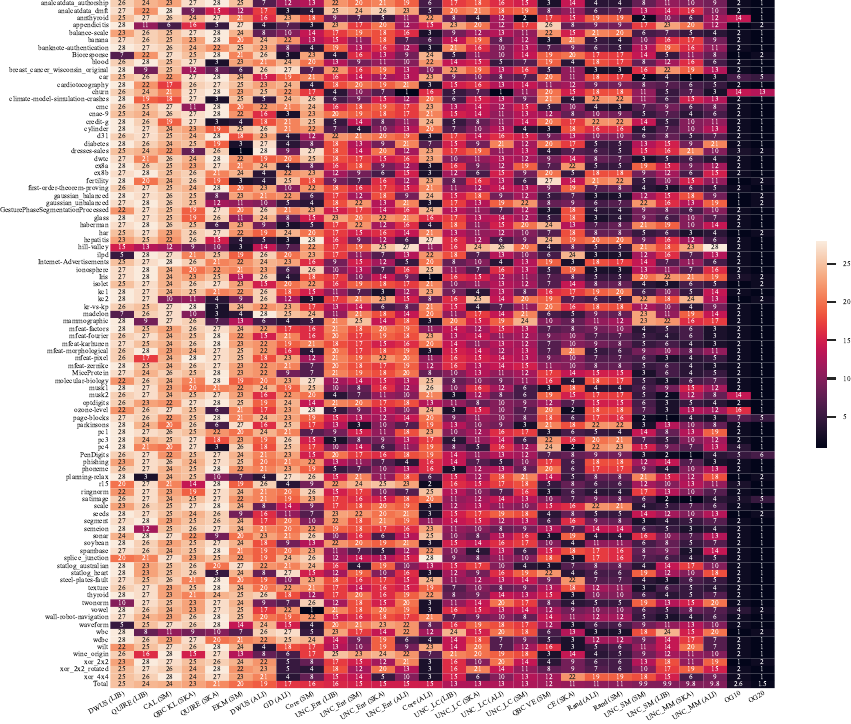}
    \caption{Final Leaderboard of all datasets and strategies, full mean of class weighted F1-score, normalized per dataset for comparability, and interpolated with 0\% for missing values (due to timeout or implementation errors)}
    \label{fig:final_leaderboard}
\end{figure*}

Given our extensive empirical results, we dare to give a non-nuanced general answer to the question of which \gls{AL} strategy works best in most scenarios, as well as showing the complete leaderboard of all datasets and parameter combinations for our entire grid in Fig.~\ref{fig:final_leaderboard}. Clearly, the margin-based \gls{AL} strategies MM and SM work best overall. These are also the only strategies able to outperform the random baseline strategies. Close behind are the Cost Embedding strategy CE, QBC in the vote-entropy (VE) variant, which performed tremendously better than the Kullback-Leibler divergence (KL) variant and the Least-Confidence (LC) and Entropy (Ent) uncertainty-variants.

It can also be clearly seen that most of the strategies implemented in several variants end up at almost identical ranks, with Core, DWUS, and QUIRE being the exceptions. Even though not shown in the plot because we use ranks, we can report that for most datasets, the differences between the strategies are not very large and smaller than 1 percent.

Also, no strategy performs consistently well on all datasets, with many having a few datasets where they perform very well and others poorly. Even the awful performing strategies like DWUS implemented in LIB or QUIRE have several datasets that rank high. One of the main arguments for the diversity-based strategy is the inability of purely uncertainty-based strategies to be unable to come out of local minima for XOR-like datasets~\cite{konyushkovaLearningActiveLearning2017, gonsiorComparingImprovingActive2024}. It can be seen that purely diversity-based strategies Coreset (Core) and Graph Density (GD) perform much better on several datasets like xor 2x2 or r15. This shows that there is indeed not a single \gls{AL} strategy that is the best choice in every scenario and explains why there is such a vast discrepancy in \gls{AL} studies regarding the best-performing strategy. 
\label{sec:leaderboard}
\section{Summarized findings}
After first defining a vast hyperparameter grid with a diverse set of hyperparameters, second performing \gls{AL} experiments for every hyperparameter parameter combination included in our grid, and third analyzing the results for correlations between the hyperparameter values, our results can be summarized as follows:
The used \gls{ML} metric has a measurable influence on the outcome of the experiments, and the used aggregation metric (area-under-curve, final value, etc.) can drastically influence the results. Our recommendation is a class-weighted F1-score in combination with the complete arithmetic mean of all \gls{AL} cycles instead of e. g. an arbitrary distinction between a ramp-up and a plateau phase. As the batch size hyperparameter also has a wide variety in the correlations among the results between different hyperparameter values, we advise using at least two batch sizes, preferably a very small and a larger value. Regarding the dataset hyperparameter, we found very high variability in the results and advise using as many datasets as possible, preferably from a well-known benchmark dataset collection. The used learner model hyperparameter only slightly changes the results and can, therefore, based on our knowledge, safely be fixed to a single, accepted domain-fitting model such as a multi-layer perceptron.
Our findings about the variance between different implementations of the same strategy and a surprisingly high correlation among strategies in the same framework, compared to strategies implemented in other frameworks, were somewhat surprising. This leads us to conclude that special care should be taken in implementing \gls{AL} strategies, as small implementation details can already have a significant influence on the outcome. Apart from that, we could find correlations among uncertainty-based strategies and among diversity-based strategies.
Multiple train-test-splits and start-point-sets seem to have an insignificant influence if (!) the overall hyperparameter grid is big enough, such as ours. The most important takeaway, in the end, is that given a hyperparameter grid with much variety among the combinations, around 4,000 randomly drawn hyperparameter combinations already show nearly the same results as the complete grid. Thereby, huge computational effort can be saved while still getting the same results as for the complete parameter grid.
Our last finding is the importance of the runtime of \gls{AL} strategies, which, given the human-in-the-loop nature of \gls{AL}, is an aspect that should not be overlooked when developing new \gls{AL} strategies.
\section{Conclusion}
We went out into this research endeavor to understand why the results of \gls{AL} research fluctuate so much between similar papers. The results of \gls{AL} research papers are often not reproducible and contradictory. Our central hypothesis is that \gls{AL} depends on many hyperparameters, yet no extensive study has investigated the cross-correlations between different hyperparameters. We have, therefore, first gathered all possible hyperparameters and carefully designed a big hyperparameter grid containing as much variety among the parameters as possible for real-world use-case scenarios. After evaluating all hyperparameter combinations on a high-performance computing cluster, we analyzed the results and got many interesting and surprising insights into the \gls{AL} research process. We outlined the dangers of using some hyperparameter values, such as different aggregation metrics other than the full arithmetic mean, or the big difference in different implementations by various authors of the same \gls{AL} strategy.

Our hopefully most significant contribution to the \gls{AL} research field is that in the end, it does not matter which hyperparameters you evaluate. The results will be reproducible and comparable to other works if you sample a few hyperparameter combinations from an extensive hyperparameter grid.
\section{Appendix}

\section*{Acknowledgments}
The authors gratefully acknowledge the computing time made available to them on the high-performance computer at the NHR Center of TU Dresden. This center is jointly supported by the Federal Ministry of Education and Research and the state governments participating in the NHR (\url{www.nhr-verein.de/unsere-partner}).
\section{References Section}

\bibliographystyle{IEEEtran}
\bibliography{zotero_updated}
\end{document}